Fall 11-18-2016

# Human Attention Detection Using AM-FM Representations

WENJING SHI

_Candidate_

_Department_

This thesis is approved, and it is acceptable in quality and form for publication:

_Approved by the Thesis Committee:_

\_\_\_\_\_\_\_\_\_\_\_\_\_\_\_\_\_\_\_\_\_\_\_\_\_\_\_\_\_\_\_\_\_\_\_\_\_\_\_\_\_\_\_\_\_\_\_\_\_\_\_ , Chairperson

# Human Attention Detection Using AM-FM Representations

by

## Wenjing Shi

B.S., Communication Engineering, Tianjin University of Commerce, 2014

THESIS

Submitted in Partial Fulfillment of the
Requirements for the Degree of

Master of Science
Electrical Engineering

The University of New Mexico

Albuquerque, New Mexico

December, 2016

# Dedication

*To my parents for their love and supporting me all the way.*
*To my family, aunt Wang and uncle Zhang. They provide lots of assistance when I*
*am in the U.S. so that I can focus on my academic study more easily.*

# Acknowledgments

I would first like to thank my thesis advisor Prof. Marios S. Pattichis. The results described in this thesis were accomplished with his participation and guidance. I would also like to thank my committee members, Prof. Sylvia Celedon-Pattichis, Prof. Ramiro Jordan, and Dr. Sergio Murillo, for their dedication and feedback. I also thank the assistant director, Venkatesh Jatla. He provided very valuable comments on this thesis. Finally, I would like to thank all my lab mates for their continued support.

# Human Attention Detection Using AM-FM Representations

by

## Wenjing Shi

B.S., Communication Engineering, Tianjin University of Commerce, 2014

M.S., Electrical Engineering, University of New Mexico, 2016

## Abstract

Human activity detection from digital videos presents many challenges to the computer vision and image processing communities. Recently, many methods have been developed to detect human activities with varying degree of success. Yet, the general human activity detection problem remains very challenging, especially when the methods need to work "in the wild" (e.g., without having precise control over the imaging geometry). The thesis explores phase-based solutions for (i) detecting faces, (ii) back of the heads, (iii) joint detection of faces and back of the heads, and (iv) whether the head is looking to the left or the right, using standard video cameras without any control on the imaging geometry. The proposed phase-based approach is based on the development of simple and robust methods that relie on the use of Amplitude Modulation - Frequency Modulation (AM-FM) models.

The approach is validated using video frames extracted from the Advancing Out-of-school Learning in Mathematics and Engineering (AOLME) project. The dataset consisted of 13,265 images from ten students looking at the camera, and 6,122 images from five students looking away from the camera. For the students facing the camera, the method was able to correctly classify 97.1% of them looking to the left and 95.9% of them looking to the right. For the students facing the back of the camera, the method was able to correctly classify 87.6% of them looking to the left and 93.3% of them looking to the right. The results indicate that AM-FM based methods hold great promise for analyzing human activity videos.

# Contents

*Contents*

# List of Figures

# List of Tables

# Chapter 1

# Introduction

Human activity analysis has advanced significantly over the last couple of years. Methods have been developed to detect, recognize humans' faces, bodies, and activities from images and videos. Basic underlying techniques include optical flow and shape template construction and matching. Within this larger context, the current thesis is focused on human attention detection. The goal is to determine whether humans are paying attention to an object or a human located to their right or left.

Human attention detection is a complicated by the need to robustly detect the positions of the different humans in an image. Then, pose estimation requires that we develop methods that understand basic facial expressions or body language.

Generally, images collected in controlled environments are much easier to analyze than ones collected "in the wild". In uncontrolled environments, lots of elements (such as light, viewing angles, resolution, etc.) will have an impact on detection results. Yet, for most practical applications, there is strong interest in developing methods for images detected in uncontrolled environments.

## 1.1 Motivation

The primary motivation of this thesis is to develop methods to detect human attention. The images were collected in an uncontrolled environments, as part of the Advancing Out-of-school Learning in Mathematics and Engineering (AOLME) project. The videos that were taken "in the wild" without full control of the camera angles.

Fig. 1.1 provides several image samples from different scenes. Students in these samples sit around the table randomly, and they either face the camera or have their back to the camera. This kind of situation also causes issues associated with non-uniform illumination.

This study will focus on the determination of students' attention. For example, if two of them have the same head direction, the assumption is that they are looking at the same thing. Otherwise, they may be talking to each other or be interested in different things which can be determined by detecting their positions in the image.

## 1.2 Thesis Statement

It is difficult to determine where people look based on AOLME videos because they were taken in uncontrolled environments. As a result, the thesis claims that (i) a phase-based method can be developed that is insensitive to illumination variations, and (ii) that the proposed method will work for human images taken from different camera angles.

The basic idea is to use multiscale Amplitude Modulation - Frequency Modulation (AM-FM) models which can provide very effective representations of the image. Then, attention detection is limited to determining whether people are looking to their right or left. On the other hand, a major innovation of the current thesis is the development of an AM-FM method for detecting human attention for students facing away from the camera (applied to girls). In all cases, AM-FM representations can clearly capture the underlying face and head textures and patterns.

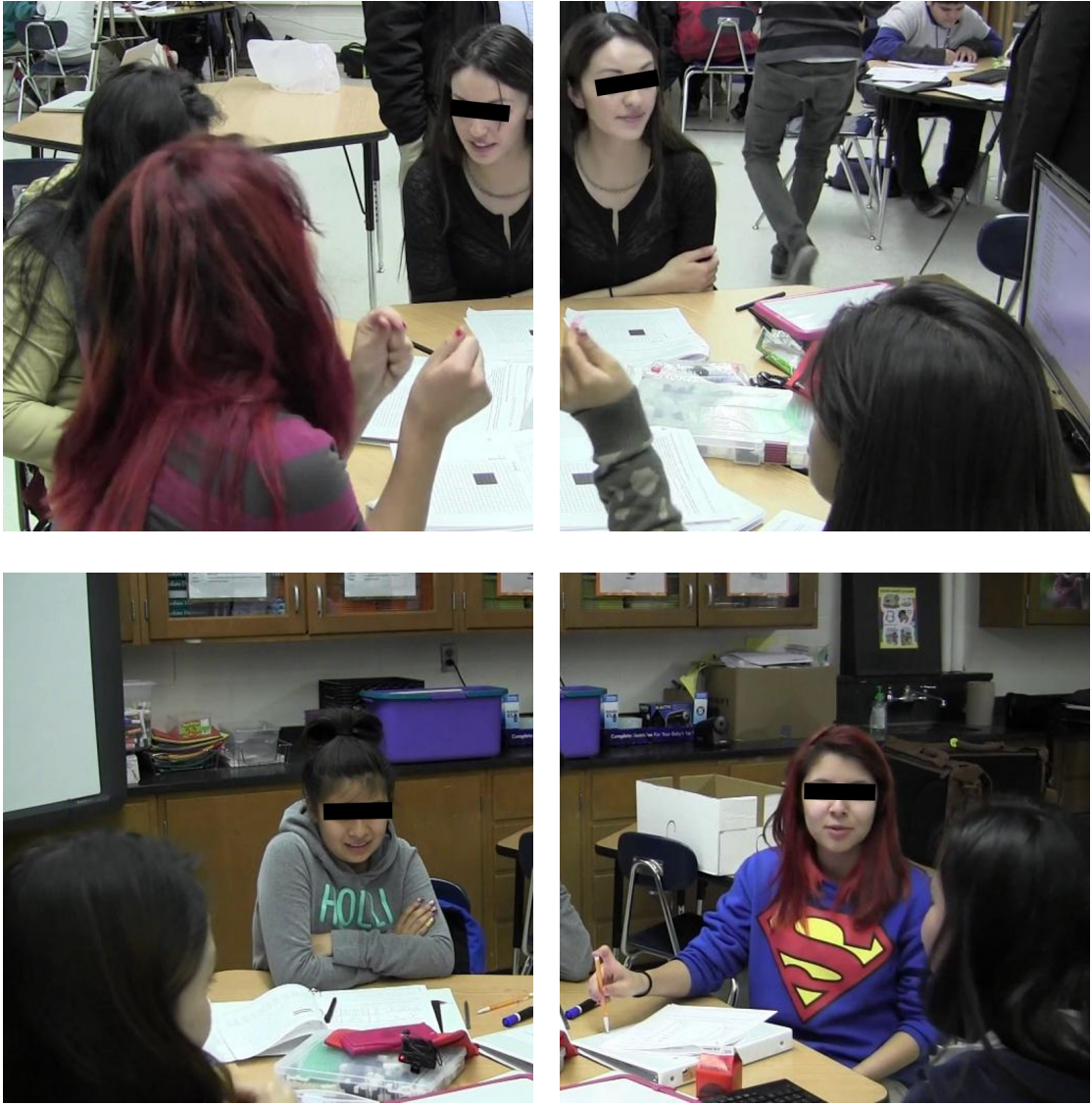

Figure 1.1: Four image samples from AOLME videos

## 1.3   Contributions

The contributions of this thesis includes:

- A phase-based method that clearly detect eyes, mouth, nose, eyebrows, laugh lines, or hair based on AM-FM representations. The extracted features are then used for subsequent processing.

- A new method for detecting the back of the head. In combination with face detection, the new method enables a new approach for detecting whether the students were looking to the left or right.

## 1.4   Thesis Overview

Chapter 2 presents the background for AM-FM representation models, Gabor filter, face recognition, and pose estimation. Chapter 3 describes the design and implementation of a daisy petal Gabor filterbank. Chapter 4 provides the implemented methodology for face and back of head direction detection. The results are presented and discussed in Chapter 5. Conclusion and future work are summarized in Chapter 6.

# Chapter 2

# Background

## 2.1 AM-FM Demodulation

Over the years, a variety of AM-FM demodulation methods have been developed. AM-FM demodulation examples include Hilbert-based approaches, variable spacing local linear spacing, Quasi-Eigenfunction Approximation (QEA) methods, and Quasi-local method (QLM) methods as described in [2] and Table 2.1. The authors in [2] also developed multiscale methods. Multiscale AM-FM methods have found many applications in biomedical image analysis as summarized in Table 2.1. An advantage of multiscale methods is that they can capture image patterns that can serve as important biomarkers or features but cannot be observed by the unaided eye. More generally, frequency modulation can provide effective methods for non-stationary image analysis. Ideal examples include fingerprint image analysis and tree ring image analysis. A summary of different applications is given in Table 2.1. Other AM-FM applications include ultrasound images [3] [4] and brain magnetic resonance images[5] texture analyses; diabetic retinopathy lesion [6] [7] and exudates in the macula detection [8]; surface electromyographic signals classification [9]; a tool for transform-domain energy compaction of broadband signals [10] and a novel class of multidimensional orthogonal FM transforms [11]construction; foveated video compression [12] and quality assessment [13], etc. In this thesis, we will use a new Gabor filterbank and relie on the basic QEA method for AM-FM demodulation. Fig. 2.1 shows the basic approach.

## 2.1.1 AM-FM Representations for Images

Assume an image $I(x,y)$ is a function of a vector of spatial coordinates $(x,y)$. Here, we are interested in describing the image as a sum of AM-FM components. For image analysis purposes, we are interested in the instantaneous amplitude (IA) $A_n(x,y)$ and instantaneous phase (IP) $\phi_n(x,y)$ components. We write:

$$I(x,y) = \sum_{n=1}^{N} I_n(x,y) = \sum_{n=1}^{N} A_n(x,y) \cos[\phi_n(x,y)]. \tag{2.1}$$

To estimate the IA and IP components, we compute the extended analytic signal with a 2D discrete Hilbert transform as given by:

$$I_{AS}(x,y) = I(x,y) + j\mathcal{H}[I(x,y)] \tag{2.2}$$

where

$$\mathcal{H}[I(x,y)] = \frac{1}{\pi x} * I(x,y). \tag{2.3}$$

After filtering the resulting image using a Gabor filterbank, we have the following approximation for the ouput of the n-th channel:

$$I_{n\_AS}(x,y) \approx A_n(x,y) \exp[j\phi_n(x,y)]. \tag{2.4}$$

Then, the amplitude and the phase components can be estimated using:

$$A_n(x,y) = |I_{n\_AS}(x,y)| \tag{2.5}$$

and

$$\phi_n\left(x, y\right) = \arctan\left[\frac{\text{imag}\left(I_{n\_AS}\left(x, y\right)\right)}{\text{real}\left(I_{n\_AS}\left(x, y\right)\right)}\right]. \tag{2.6}$$

Using the estimates of IA and IP, an image can be reconstructed based on the AM-FM decomposition by (2.1). The $\cos\left[\phi_n\left(x, y\right)\right]$ define the frequency-modulated (FM) components, which also describe strong variations in image intensity.

The AM-FM representation of (2.1) can also be used to describe multi-scale decompositions. In this case, the index $n = 1, 2, ..., M$ represents different scales [2]. Different scales are then defined by specifying the filters for each scale. At each pixel, for the filters belonging to the corresponding scale, the AM-FM component is reconstructed by selecting the AM-FM channel estimates with the largest IA. The approach recovers the underlying image texture which cannot be observed with the unaided eye.

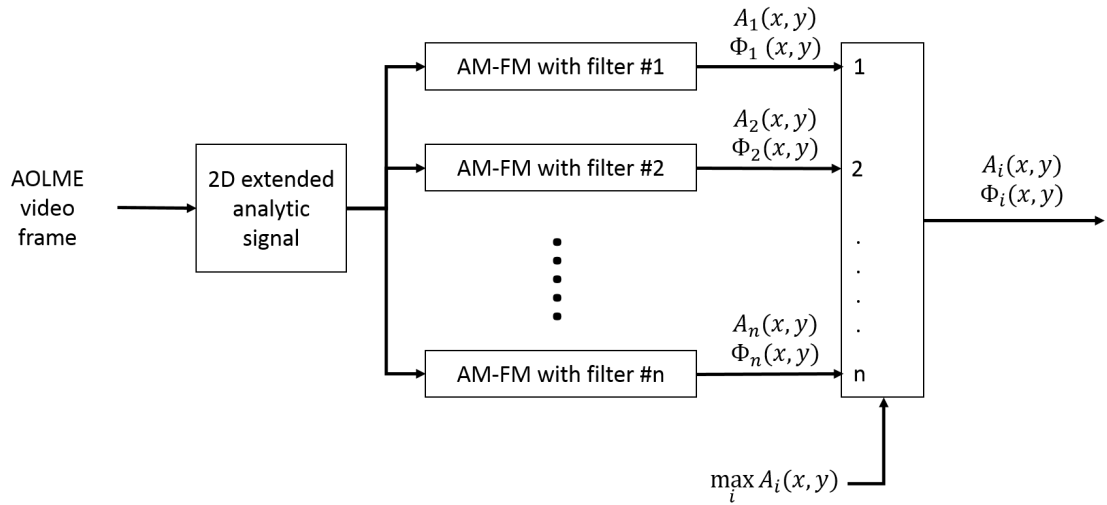

Figure 2.1: AM-FM demodulation.

## 2.1.2 Gabor Filters

As shown in Fig. 2.1, there are several different filter types that can be used in implemented for estimating multiscale AM-FM models. In this thesis, we focus on the use of Gabor filters which have shown great promise in human image analysis. The current thesis attempts to reproduce the daisy petal Gabor filterbank described in [21] and uses this approach for image analysis. The motivation for the daisy-petal method comes from its directional properties and its nice coverage of the 2D frequency plane with few gaps.

A Gabor filter is given by:

$$G(x,y) = \frac{1}{2\pi\gamma\sigma^2} \exp\left(-\frac{\left(\frac{1}{\gamma}x'\right)^2 + y'^2}{2\sigma^2}\right) \exp\left(j\left(2\pi F x'\right)\right) \tag{2.7}$$

where the real part is:

$$G_r(x,y) = \frac{1}{2\pi\gamma\sigma^2} \exp\left(-\frac{\left(\frac{1}{\gamma}x'\right)^2 + y'^2}{2\sigma^2}\right) \cos\left(2\pi F x'\right), \tag{2.8}$$

and the imaginary part is:

$$G_i(x,y) = \frac{1}{2\pi\gamma\sigma^2} \exp\left(-\frac{\left(\frac{1}{\gamma}x'\right)^2 + y'^2}{2\sigma^2}\right) \sin\left(2\pi F x'\right), \tag{2.9}$$

and the rotated coordinates are given by:

$$x' = x\cos\theta + y\sin\theta, \quad \text{and} \tag{2.10}$$

$$y' = -x\sin\theta + y\cos\theta. \tag{2.11}$$

Here, $\sigma$ is the standard deviation of the Gaussian envelope and it determines the size of a filter kernel. Fig. 2.2 shows several sample images with different $\sigma$ values. $\gamma$ is the spatial aspect ratio and it sets the aspect ratio of the kernel. Fig. 2.3 presents four kernels with various $\gamma$. In this study, $\gamma = 0.5$. $F$ is radial center frequency which is measured in cycles/image, and $\theta$ is the orientation, which is measured by degrees or radians from the u-axis. Fig. 2.4 shows four Gabor filter kernel samples with different orientations. Furthermore, $F$ and $\theta$ are defined using

$$F = \sqrt{u^2 + v^2}, \tag{2.12}$$

and

$$\theta = \arctan\left(\frac{v}{u}\right), \tag{2.13}$$

where $(u, v)$ is the center frequency,

$$u = L\cos\left(2\pi\frac{\texttt{Ang}}{360}\right)$$
$$v = L\sin\left(2\pi\frac{\texttt{Ang}}{360}\right)$$

and $\texttt{Ang}$ is the angle of the filter measured in degrees.

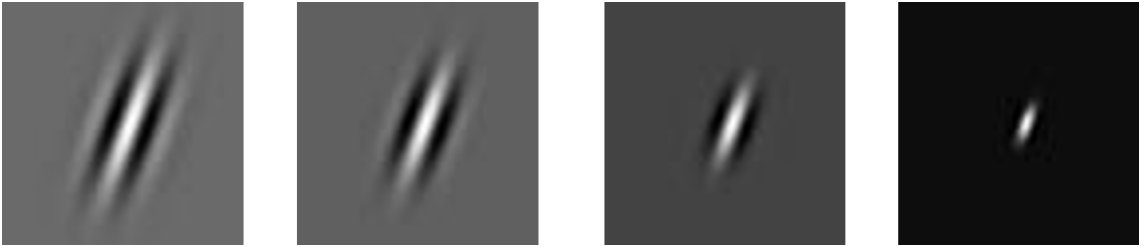

Figure 2.2: Gabor filter kernels with different sizes: $\sigma$=11, 9, 6 and 3. Here, $\texttt{Ang} = 20.25°$ and $F = 27/128 \cdot \pi$.

The frequency and orientation of a Gabor filter can be changed. Gabor filters have been considered in different applications. For example, a face representation based on Local Gabor Binary Pattern Histogram Sequence (LGBPHS) was proposed in [22]. This method was motivated as an approach to reduce lighting artifacts, face expression variability, and aging artifacts.

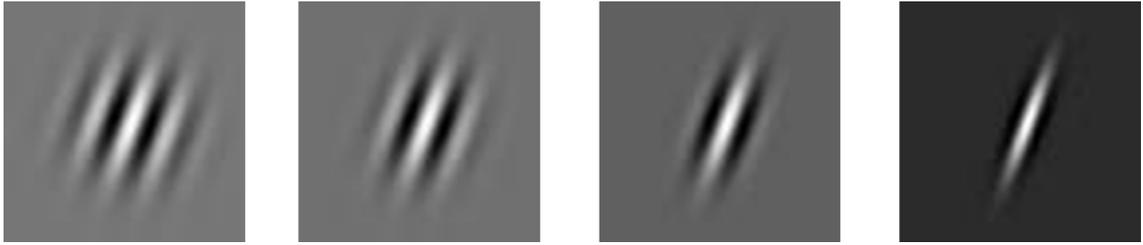

Figure 2.3: Gabor filter kernels with different aspect ratios: $\gamma$=1, 0.75, 0.5 and 0.25. Here, $\sigma = 9$.

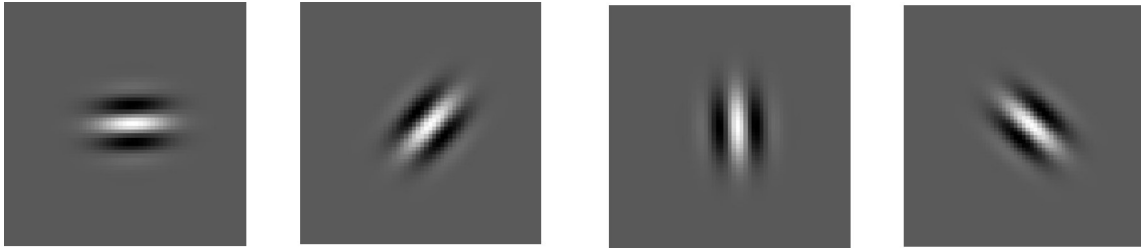

Figure 2.4: Gabor filter kernels with different orientations: $\theta = 0°, 45°, 90°$ and $135°$, $F = 27/128 \cdot \pi, \texttt{Ang} = 90°, 45°, 0°, 135°, \sigma = 5$.

## 2.2 Face Detection

Face detection is often used to locate a person's position. Face recognition methods were recently classified in [23] as: 1) intensity image based; 2) video based; and 3) using other sensory data such as 3D information or infra-red imagery. Most methods are image intensity based. This thesis will use both intensity and texture based methods derived from single images.

Faces have distinct color and texture information. For face color region detection, accuracy can be greatly reduced due to artifacts resulting from non-uniform illumination. To avoid this problem, methods based on the unique locations of facial features have been proposed. Table 2.2 lists common datasets used for face detection and Table 2.3 shows some of the latest approaches for face detection.

From Table 2.3, we can see that a phase based method to detect faces was also used in [24]. In this study, for this research, due to the uniqueness of AOLME datasets, not only face detection, but also back of the head detection will need to be considered to find the position of the head. For this method, especially for the back of the head detection, no previous approach considered a phase based method.

## 2.3   Pose Estimation

Human pose estimation can be very challenging. Some researchers used specialized hardware to obtain more pose features, like RGB-D camera and different sensors [35]. Such methods are more suitable for whole body pose estimation than head poses estimation. 3D models are also currently being researched. For pose estimation in videos, methods based on the connection of multiple frames have been proposed. In typical cases, analytic or geometric methods, genetic algorithm methods, and learning based methods are applied to estimate pose.

Tables 2.4 and 2.5 list the common datasets and some recent methods for human pose estimation, respectively. The summary focuses on head pose estimation and the application of analytic geometry methods.

In this thesis, AM-FM decompositions were used to represent images with 54 daisy petal Gabor filters for filtering. Based on this filterbank, more features were captured than the standard methods described in this chapter. The additional features benefited subsequent analysis. For face detection, the thesis includes both color region detection and KNN classification, resulting in improved detections. For pose estimation, the thesis focuses on head poses, and applies the head pose method based on the distribution of the different face features.

Table 2.1: Recent AM-FM Methods

| Author | Filterbank | AM-FM Demodulation Method | Application |
|---|---|---|---|
| Agurto, et al. [14] | Dyadic filterbank | VS-LLP | Neovascularization in the optic disc (NVD) detection |
| Pattichis, et al. [15] | Gabor filterbank | Teager-Kaiser operator based | Frequency modulation theory demonstrated on woodgrain and fingerprint images. |
| Ramachandran, et al. [16] | A variable frequency spacing filter bank integer-based Savitzky-Golay filter | Hilbert-based AM-FM | Tree growth analysis |
| Pattichis, et al. [17] | Gabor channel filters | QEA | Fingerprint classification |
| Pattichis, et al. [18] | A collection of Gabor channel filters | QEA | Electron micrographs of skeletal muscle segmentation |
| Loizou, et al.[19] | Dyadic 2D filterbank | VS-LLP | Segmentation and classification in the carotid artery |
| Belaid, et al.[20] | Quadrature filters | Monogenic Signal | Segmentation of ultrasound images |

Table 2.2: Common datasets used for face recognition.

| Title | Description | URL |
|---|---|---|
| LFW [25] | More than 13,000 face images collected from the web with large variations in pose, age, expression, illumination, etc. | `http://vis-www.cs.` `umass.edu/lfw/` |
| YTF [26] | 3,425 videos of 1,595 different people. All the videos were downloaded from You tube. An average of 2.15 videos are available for each subject. The shortest clip duration is 48 frames, the longest clip is 6,070 frames, and the average length of a video clip is 181.3 frames. | `http://www.cs.tau.ac.` `il/~wolf/ytfaces/` |
| Multi-PIE [27] | More than 750,000 images of 337 people recorded in up to four sessions over five months. | `http:` `//www.flintbox.com/` `public/project/4742/` |
| CASIA-WebFace [28] | 10,575 subjects and 494,414 images. | `http://www.cbsr.ia.ac.` `cn/english/` `CASIA-WebFace-Database.` `html` |
| CACD [29] | 163,446 images from 2,000 celebrities collected from the Internet. | `http://bcsiriuschen.` `github.io/CARC/` |
| AFLW [30] | The database contains about 25k annotated faces in real-world images. Of these faces 59% are tagged as female, 41% are tagged as male (updated); some images contain multiple faces. | `https://lrs.icg.tugraz.` `at/research/aflw/` |

Table 2.3: Recent face detection methods.

| Author | Method | Feature | Comment |
|---|---|---|---|
| Kim. et al. [31] | Multiple color spaces | Investigated the effectiveness of facial color in face recognition with deep learning, feature fusion. | Face recognition accuracy with color face image was significantly higher than that of using grayscale images. |
| Dahmane. et al.[24] | Fourier transform, phase-context, local phase quantization, texture representation. | Texture representation using phase-context which is based on four-quadrant mask of the Fourier transform phase in local neighborhoods. | Representing face from low-resolution images under varying light, illumination and blur. |
| Zhu. et al. [32] | Using a mixture of trees with a shared pool of parts. | Model every facial landmark as a part and use global mixtures. | Effective at capturing global elastic deformation and it is easy to optimize. |
| Cui. et al.[33] | Uses robust face region descriptors | Developed a new pairwise constrained multiple metric learning (PMML) method. | Face can easily be aligned accurately with complex appearance variations or from low-quality images. |
| Schroff. et al.[34] | Presenting a system, FaceNet | Learning a mapping from face images to a compact Euclidean space where distances directly correspond to a measure of face similarity directly. | Face recognition, verification and clustering can be easily implemented using standard techniques with FaceNet embedding as feature vectors. |

Table 2.4:  Common datasets used for pose estimation.

| Title | Description | URL |
|---|---|---|
| Frames Labeled In Cinema (FLIC) [36] | 5003 image dataset from popular Hollywood movies. | `http://bensapp.github.io/flic-dataset.html` |
| Leeds Sports Pose Dataset [37] | 2000 pose annotated images of mostly sports people gathered from Flickr. | `http://www.comp.leeds.ac.uk/mat4saj/lsp.html` |
| MPII [38] | 25K images containing over 40K people with annotated body joints. | `http://human-pose.mpi-inf.mpg.de/` |
| VGG [39][40][41][42][43][44][45] | The VGG Human Pose Estimation datasets includes large video datasets annotated with human upper-body pose. | `https://www.robots.ox.ac.uk/~vgg/data/pose/` |
| MPIIGaze [46] | 213,659 images from 15 participants. The number of images collected by each participant varied from 34,745 to 1,498. | `https://www.mpi-inf.mpg.de/departments/computer-vision-and-\multimodal-computing/research/gaze-based-human-\computer-interaction/appearance-based-gaze-\estimation-in-the-wild-\mpiigaze/` |

Table 2.5: Recent pose estimation methods.

| Author | Method | Feature | Comment |
|--------|--------|---------|---------|
| Xu. et al. [35] | RGB-D camera, using two depth sensors simultaneously. | Captured the front and back of the body's movement. Wide baseline RGB-D camera calibration algorithm, Gaussian mixture model. | The reconstruction of detailed human model was greatly improved. |
| Liu. et al. [47] | Convolutional neural network, which is trained on synthetic head images. | Evaluated the method on both synthetic and real data. | Generated a realistic head pose dataset by rendering techniques, including different gender, age, race and expression. |
| Pfister. et al. [48] | ConvNet architecture. | Combined information across the multiple frames using optical flow. | Simple, direct method for regressing heat maps. Results were improved by combining the regression with optical flow and spatial fusion layers. |
| Sapp. et al. [49] | A multi-modal, decomposable model (MODEC) for articulated human pose estimation in monocular images. | Used a linear structured model, which struggled to capture the wide range of appearance present in realistic, unconstrained images. | The model provided a way to maintain the efficiency of simple, tractable models while gaining the rich modeling power of many global modes. |

# Chapter 3

# Gabor Filter Bank Design

Compared to other filters, Gabor filters have the advantage that they closely approximate filtering processes associated with the human visual system. In this thesis, we consider a filterbank based on 54 daisy petal Gabor filters based on the discussion of Chapter 2.1.2.

In Fig. 3.1, $L$ is the distance between a filter's center and the origin, and `Ang` represents the orientation of the filter. In Fig. 3.2, we show the frequency responses with two different Gabor filters.

For filter bank design, we compute the overlaps between three consecutive filters along the same direction as shown in Figs. 3.3 and 3.4. In the example, assuming a peak of $M$, the filters overlap at $0.8 \cdot M$. Based on this overlap rule, the filter shown in Fig. 3.5 was generated. Fig. 3.6 illustrates the multi-scale design. The orientation directions are uniformly sampled. A full description is given in Table 3.1. We have 54 pairs of values for parameters $L$, $\sigma$ correspond to various values of `Ang` to generate a daisy petal Gabor filter bank. Figs. 3.7 and 3.8 show the frequency response.

Fig. 3.9 shows the estimated FM component using dominant component analysis with the daisy petal Gabor filterbank. From Fig. 3.9, eyes, eyebrows, nose, mouth, hairs, even wrinkles on the clothes, can be observed. This example shows that the filterbank worked very well for these images taken "in the wild".

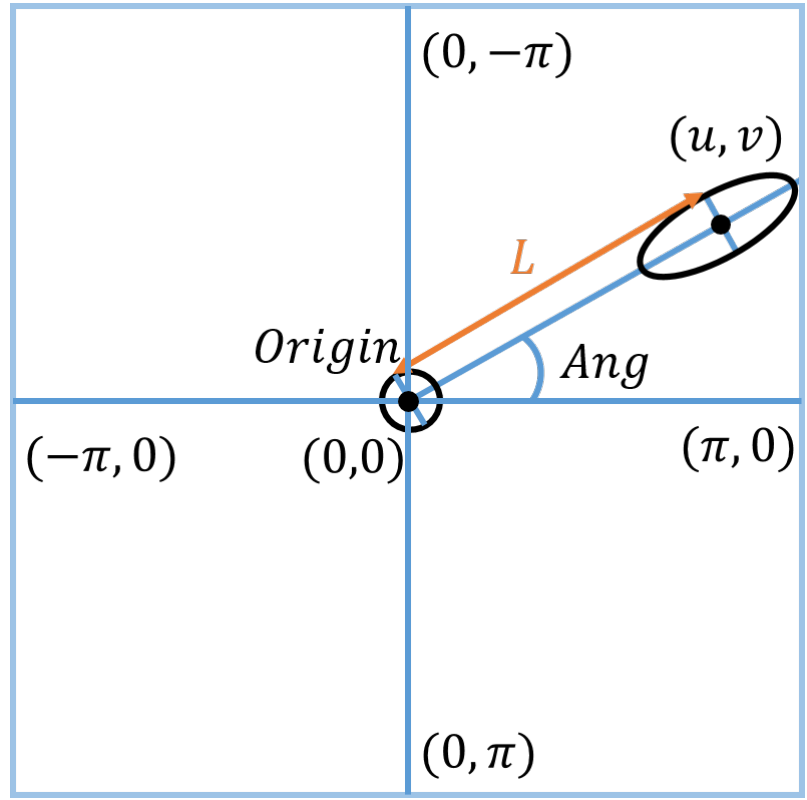

Figure 3.1: Geometry for a filter in the spatial frequency response

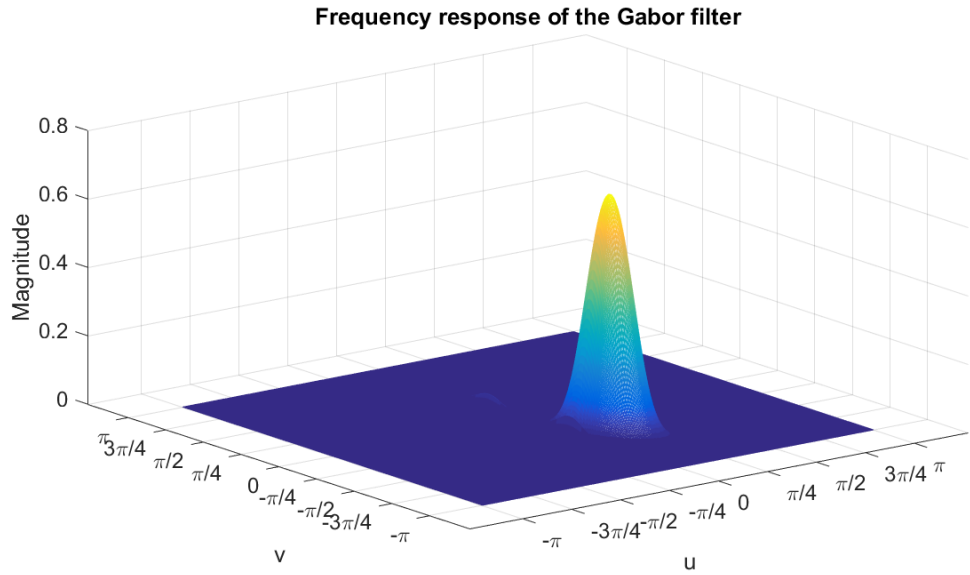

(a)

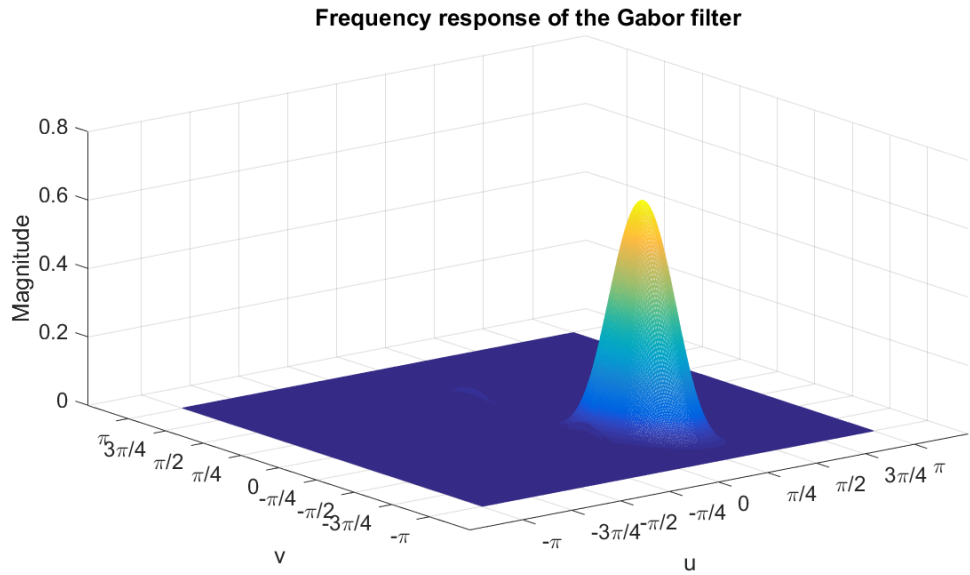

(b)

Figure 3.2: Frequency response of different Gabor filters: (a) $L = 0.2\pi$, $Ang = 20.25°$, $\sigma=7$, $\gamma=0.5$. (b) $L = 0.3\pi$, $Ang = 20.25°$, $\sigma=5$, $\gamma=0.5$.

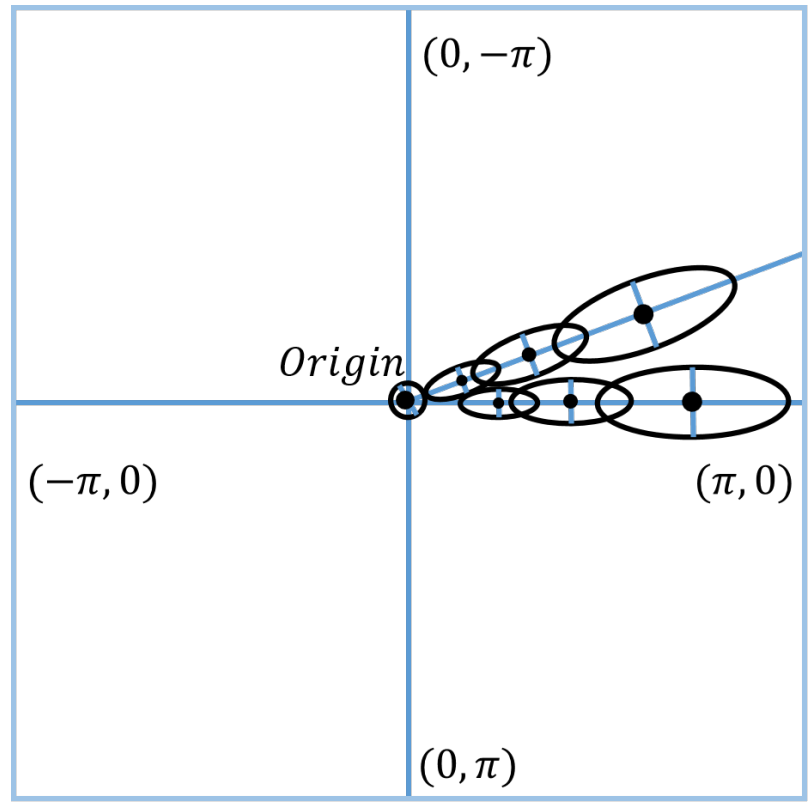

Figure 3.3: Geometry for several filters in the spatial frequency responses

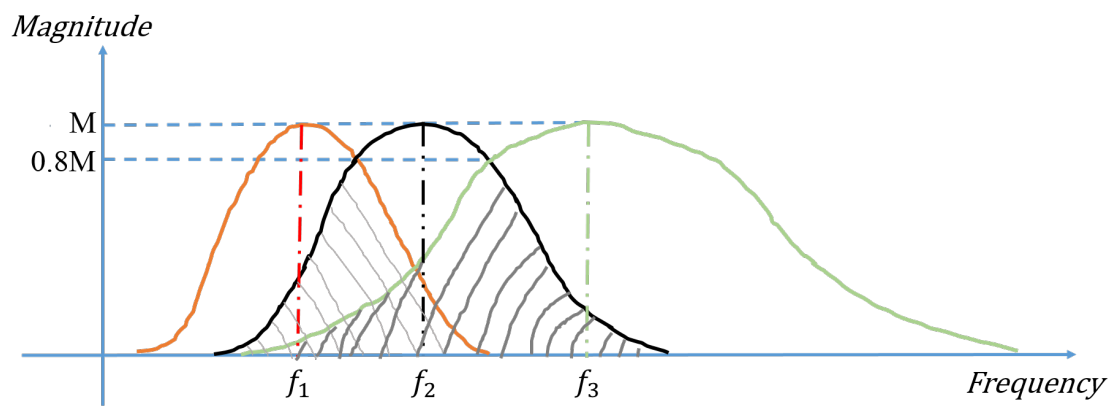

Figure 3.4: The overlaps between two filters in the spatial frequency responses

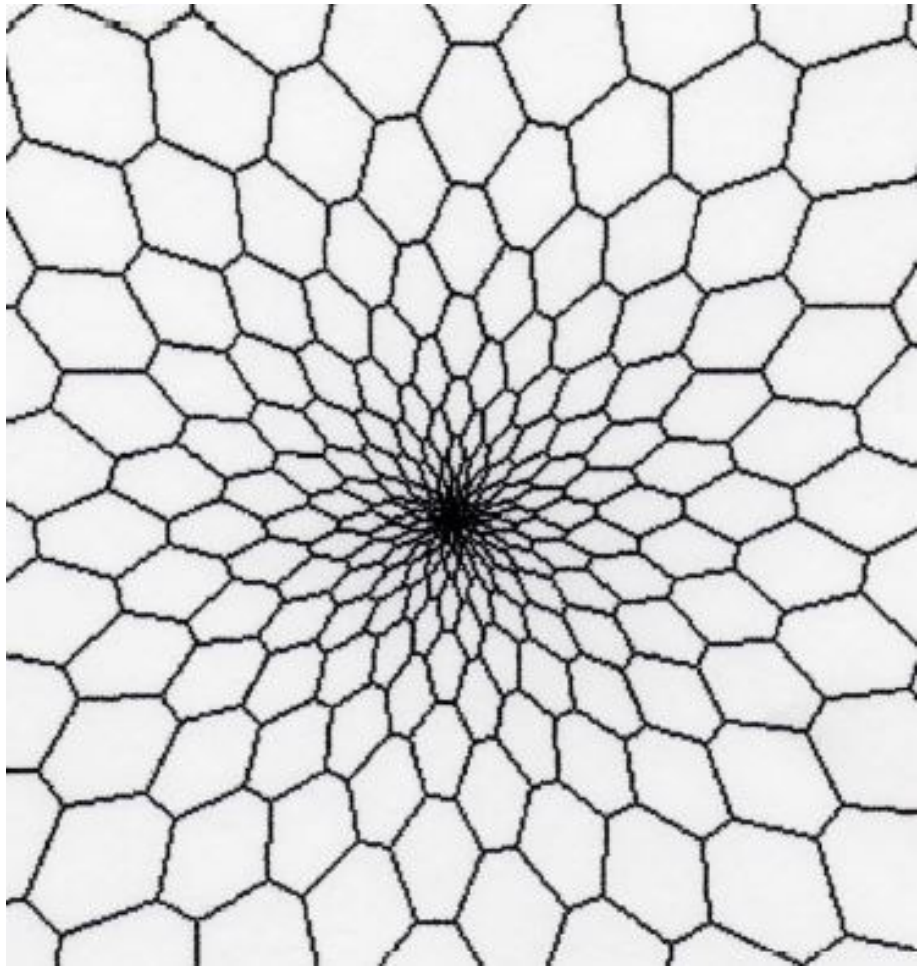

Figure 3.5: Aperiodic hexagon tiling [1]

Table 3.1: Parameters for Gabor filterbank generation

| $(L, \sigma)$ | $Ang$ | Comment |
| --- | --- | --- |
| $(0.047\pi, 11)$,<br>$(0.125\pi, 6)$,<br>$(0.242\pi, 4)$,<br>$(0.406\pi, 3)$,<br>$(0.648\pi, 2)$,<br>$(0.938\pi, 2)$ | $20.25°$<br>$65.25°$<br>$110.25°$<br>$155.25°$ | Uniform angles, the interval<br>between two angles is $45°$;<br>have intersection between<br>two filters in the same angle |
| $(0.102\pi, 7)$,<br>$(0.195\pi, 6)$,<br>$(0.313\pi, 4)$,<br>$(0.461\pi, 3)$,<br>$(0.695\pi, 2)$ | $42.75°$<br>$87.75°$<br>$133.75°$<br>$177.75°$ | Uniform angles, the interval<br>between two angles is $45°$;<br>have intersection between<br>two filters in the same angle |
| $(0.094\pi, 3)$ | $10°$<br>$32.5°$<br>$55°$<br>$77.5°$<br>$100°$<br>$122.5°$<br>$145°$<br>$167.5°$ | Uniform angles, the interval<br>between two angles is $22.5°$ |
| $(1.094\pi, 2)$ | $43.5°$<br>$133.5°$ | Uniform angles, the interval<br>between two angles is $90°$ |

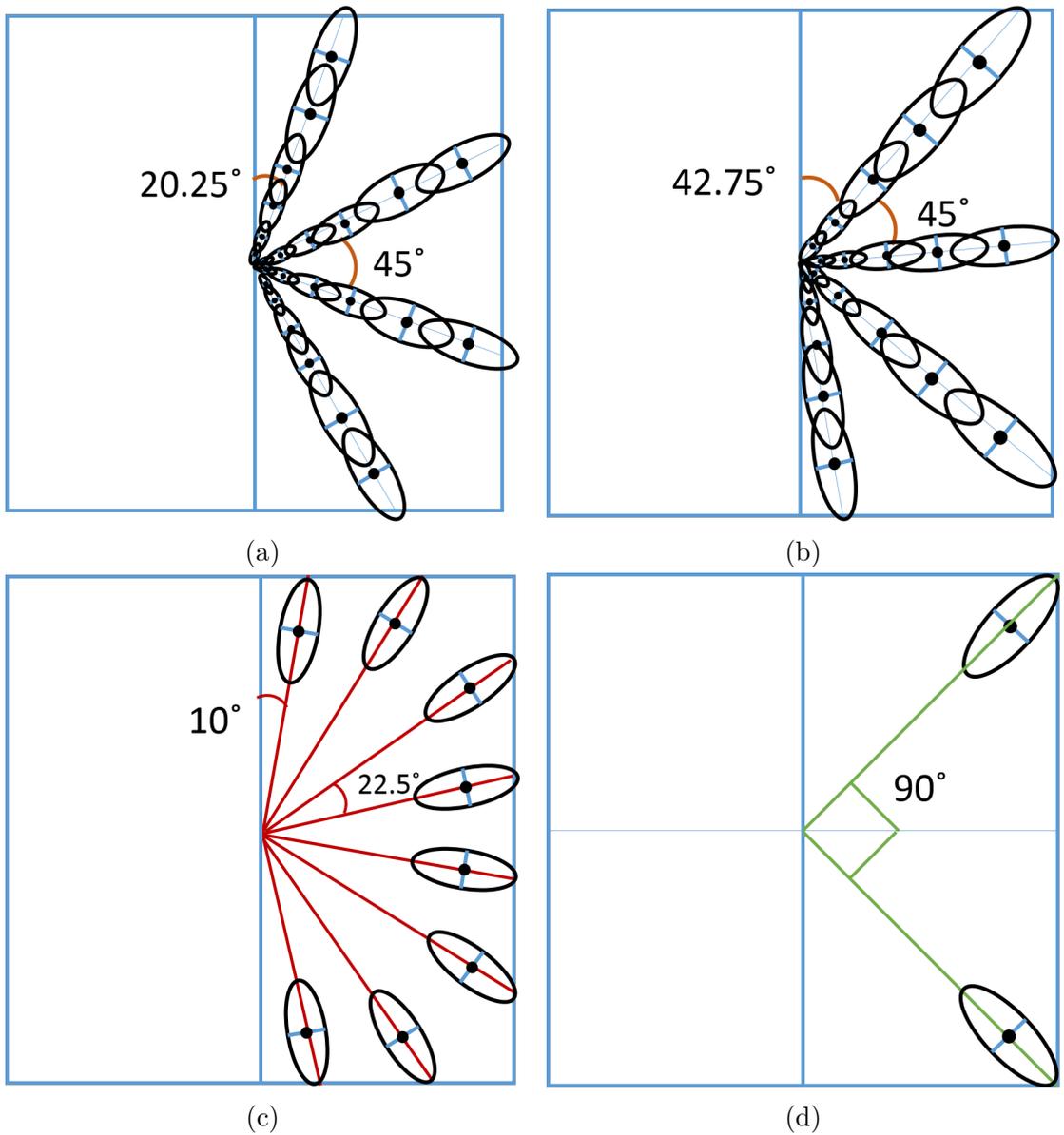

Figure 3.6: Multi-scale filterbanks design: (a) six scale filterbank. (b) five scale filterbank. (c) single scale. (d) single scale.

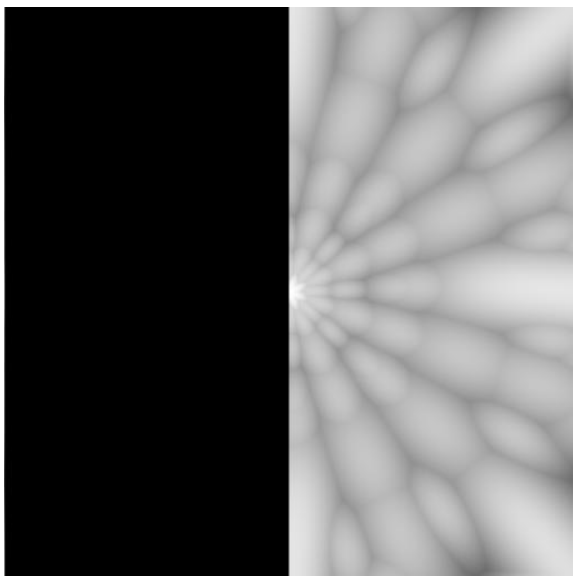

Figure 3.7: Spatial frequency response of daisy petal Gabor filter bank with 54 filters

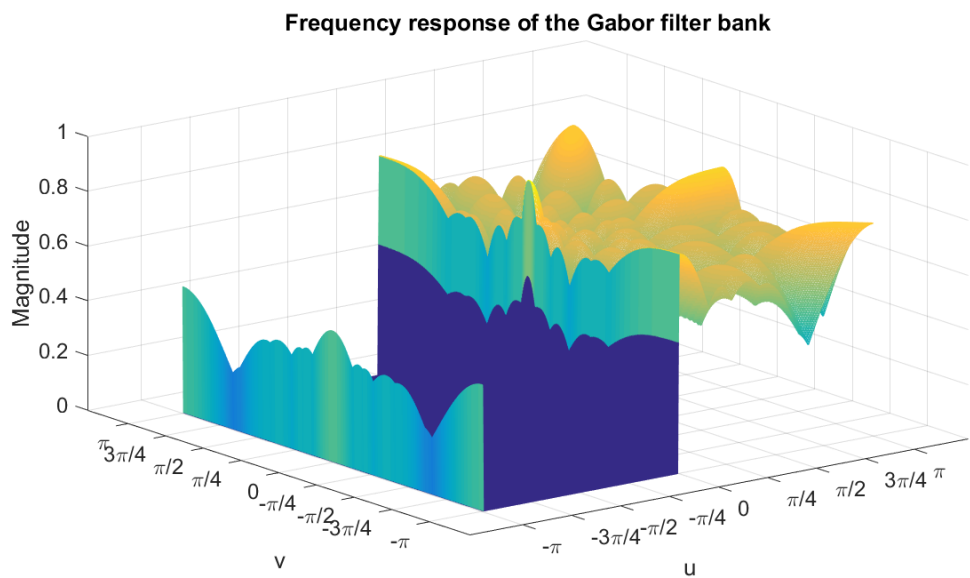

(a)

Figure 3.8: Frequency response of the Gabor filter bank

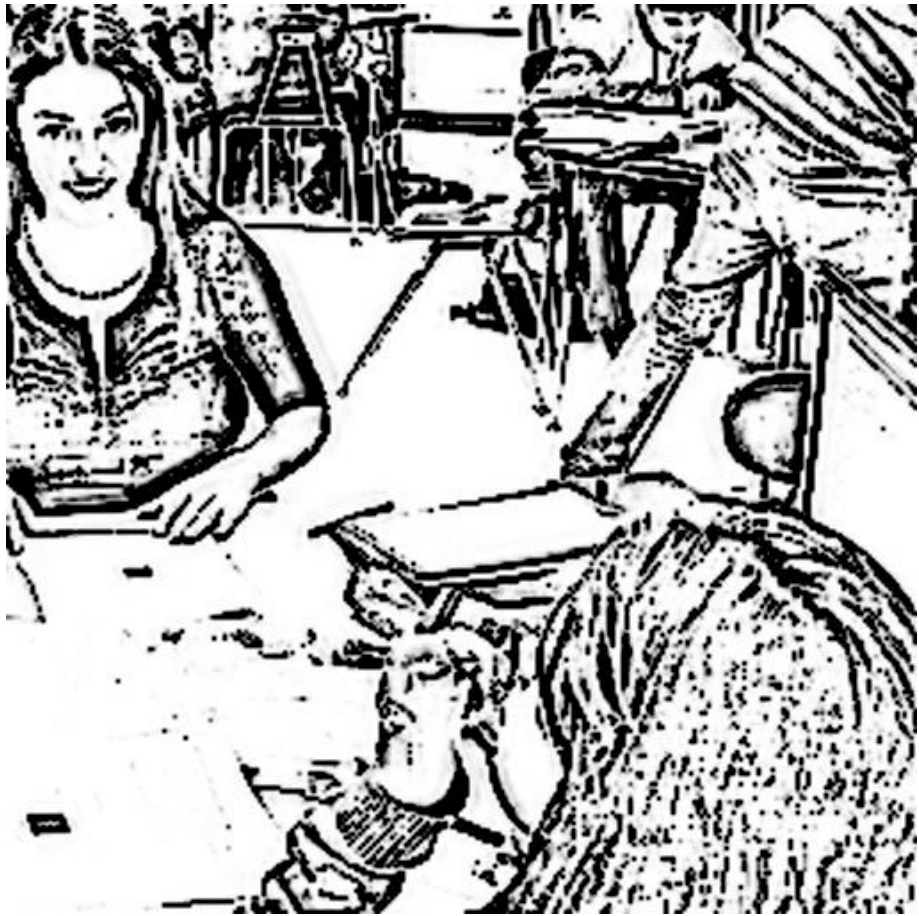

(a)

Figure 3.9: FM component with daisy petal Gabor filter bank in this work.

# Chapter 4

# Attention Detection

Human attention is determined based on head orientation. In this chapter we discuss techniques and methods to detect if a person is looking the right or left of the camera. Before determining the direction, we need to be able to detect head position. Due to uncontrolled nature of the videos to be processed in this thesis, we need to determine attention direction using both the face and the back of the head (see Fig. 4.2). A block diagram giving an overview of framework is given in Fig. 4.1.

## 4.1   Face Detection

Face detection is achieved by combining skin region detection with FM component based classification. Fig. 4.3 gives an overview block diagram of this detection method.

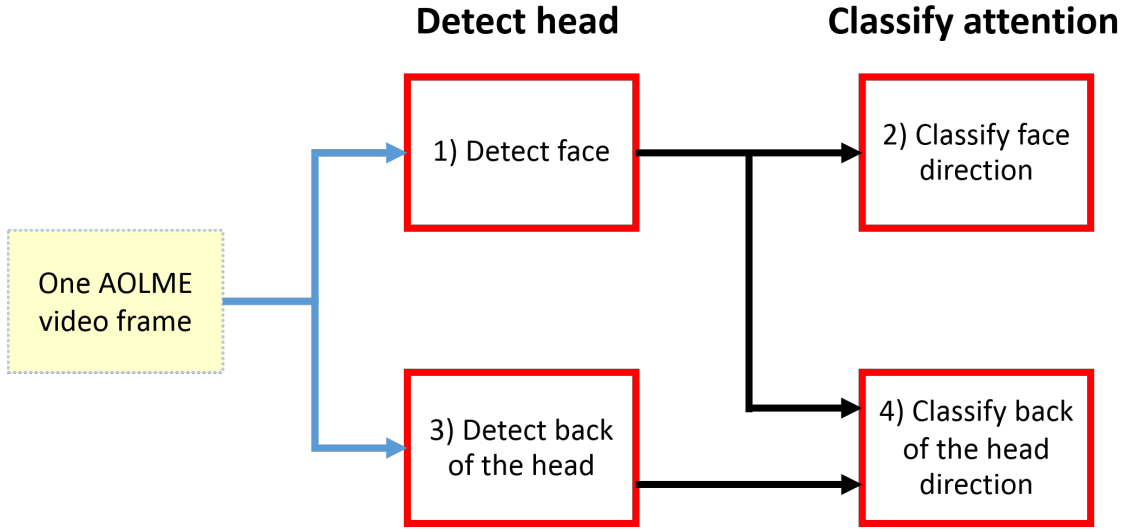

Figure 4.1: Detection method overview.

### 4.1.1 Skin Region Detection

Segmenting skin from non-skin regions requires reliable color models. For the thesis, skin detection is performed in the RGB, HSV and YCbCr color spaces. These models have been widely used in color based detection. When the input is an RGB-uint8 image, each element is an integer with goes from 0 to 255. For the YCbCr model, Y ranges from 16 to 235, while Cb and Cr take values from 16 to 240. For the HSV model, the elements are in the range of 0 to 1.

To setup the values, the thesis considered color model rules [50]: (i) the uniform daylight illumination rule, and the (ii) flashlight or daylight, lateral illumination rule. Based on these bounding rules, the parameters were manually set based on some observations. Using these models, human skin is detected effectively in AOLME videos. The four bounding rules detecting skin are given in Algorithm `DetectSkinRegions(.)` in Fig. 4.4.

The results of skin segmentation from each color model are ANDed together to generate the final skin regions. An example is given in Fig. 4.5.

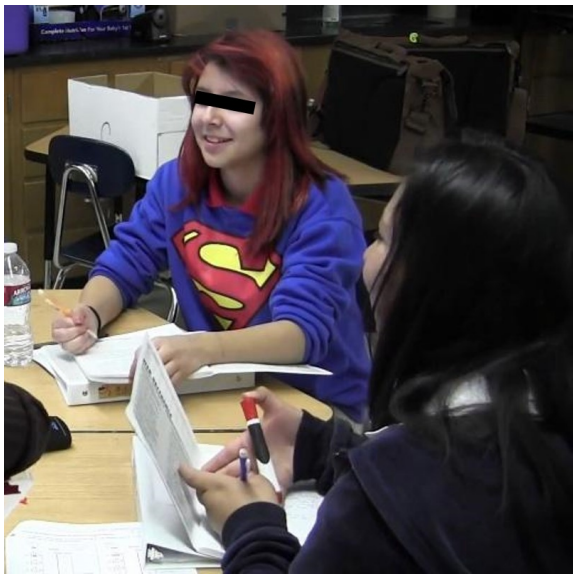

(a)

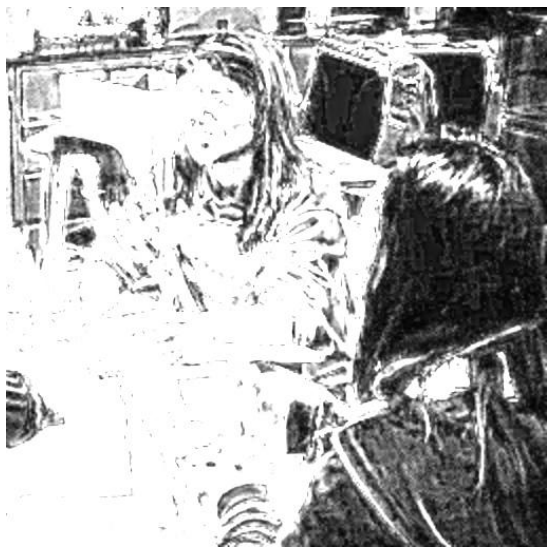

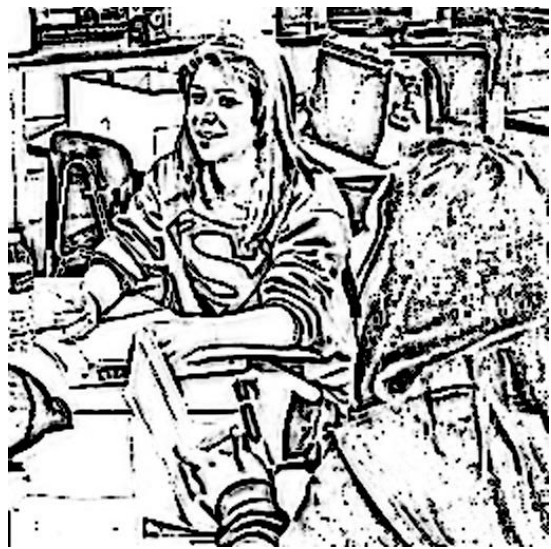

(b) (c)

Figure 4.2: A sample image example: (a) Original image. (b) AM component. (c) FM component.

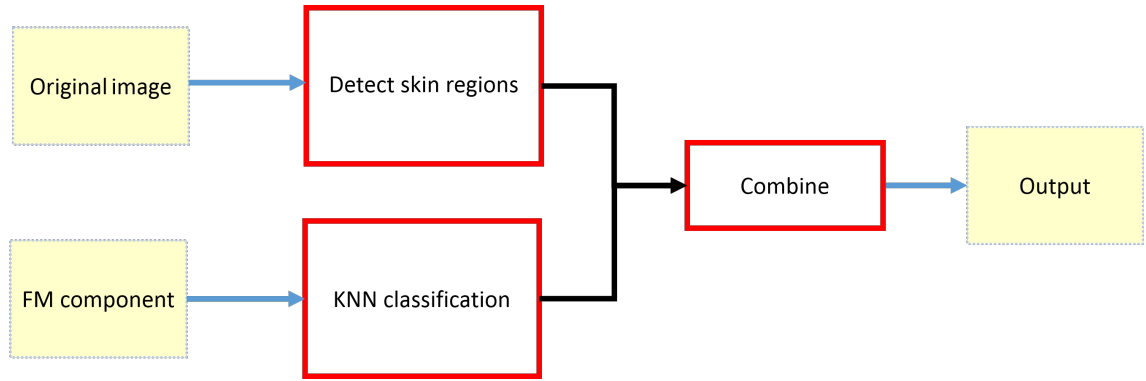

Figure 4.3: Face detection.

### 4.1.2 KNN Classification using FM images

Face detection is done for every $60 \times 60$ block with 50% overlap. For each block, a KNN classifier (K=3) is applied based on extracted FM images. Here, the training set consisted of 2442 face blocks and 2275 non-face blocks. For the final detection, the skin detection results need to be combined with the KNN classification results.

### 4.1.3 Combination

The final result is obtained by anding the results from the skin and FM classifiers (see Figs. 4.6a, 4.6a, 4.6b, 4.6c, 4.6d). The extracted face block shown in Fig. 4.7 is used for attention detection.

## 4.2 Attention Classification For Students Facing the Camera

Attention detection is based on the FM component image. The basic idea is to decompose each face into four regions and then determine attention based on the number of pixels in each segment. Thus, if more pixels are found in the right regions, the image is classified as looking to the right. A similar idea is applied for students looking the left. The method is summarized in Fig. 4.8 and an example is given in Fig. 4.10a.

As described earlier, Otsu's method [51] is used for determining the face regions. For the thesis, the method used a 256-bin image histogram. This binarized FM image is thus brought back from the face detector.

To detect the face regions we extract the largest connected component from the binarized FM image. Then, we fill-in the holes of this component to avoid any gaps [52]. The outline of the filled component is further refined by using a convex full as shown in Fig. 4.10b. From the resulting image, we compute the centroid and decompose the image into four patches (see Figs. 4.10b, 4.10c). From Fig. 4.10c, we can see several facial features (e.g., eyebrows, eyes, nose and mouth).

The four patches are illustrated in Fig. 4.9. To reduce the noise in the extracted face image, we select the 7 rows with the largest number of pixels as shown in Fig. 4.10d. From the resulting image, we compute the number of pixels in each one of the four patches. Then to detect where students are looking, we consider three classifiers: (i) lower-face classifier, (ii) upper-face classifier, and (iii) whole-face classifier. The algorithm is given as function `ClassifyFaceDirection(.)` and presented in Fig. 4.12.

## 4.3   Back of the Head Detection

To detect the back of the head, we combine the AM and FM components as shown in Fig. 4.13. In what follows, we provide a summary of each component.

We begin with the AM component (see Fig. 4.2b). In the AM component, the back of the head appears dark. Thus, the AM detector used Otsu's method to automatically determine the threshold over the whole image.

For the FM detector, refer to Fig. 4.2c. To construct a hair detector using the FM image, the basic idea is to find dark regions that are also characterized by strong vertical stripes. The basic method is illustrated in Fig. 4.15. Initially, Otsu's method is used for detecting the dark regions [51]. Then, to detect the vertical stripes, we follow a three-step approach. First, a Canny edge detector is used for determining the vertical hair lines [53]. Second, a global fill is used for enhancing the direction of each image component. Third, the resulting image is complemented so that the hair regions re-appear as dark. Here, in our example of Fig. 4.2c, we note that the global fill filters out clothes and the table, while preserving the hair.

The AM and FM detector results are ANDed together as shown in Fig. 4.16b. To detect the vertical stripes, we compute the number of pixels along each column. We then select the columns with the largest number of pixels $M = 60$ and all other columns are zeroed-out as noise. To locate the back of the head, we need a detector that finds the largest density of vertical pixels. This is done by scanning the image using $200 \times 200$ blocks, advancing each block by 1 pixel, and counts the number of pixels inside each block. The block center with the maximum number of counted pixels is then taken as the location of the back of the head (see Figs. 4.16c, 4.16d). The pseudo-code for the algorithm is given as algorithm `FindTheHighestDotDensityArea(.)` in Fig. 4.14. Lastly, the detected region is further refined by restricting it to the largest connected component in the AM image

(see Fig. 4.17).

## 4.4 Attention Classification For Students Facing Away from the Camera

To determine whether a student is looking to the left or the right, we need to combine the results from face detection and back of the head detection. The basic idea is to classify students as looking to the right if the face is detected to be on the right of the head and similarly for the left (see Fig. 4.18). Skin detection was based on the color information only (see Figs. 4.5, 4.19a). Then, we associate a face with a back of the head if they overlap. We declare an overlap when the bounding box associated with the skin detector is found to overlap with the back of the head detector. Then, to determine whether the students are looking to the left or right, we use the centroids of the bounding boxes.

Let $(x_f, y_f)$ denote the centroid for the detected face region. Then, let $(x_b, y_b)$ denote the centroid for the back of the head detector. Attention classification is performed using:

```
if xf > xb then
    Classify Right
else
    Classify Left
end if
```

```
%%%%%%%%%%%%%%%%%%%%%%%%%%%%%%%%%%%%%%%%%%%%%%%%%%%%%%%%
function DetectSkinRegions (R, G, B, H, S, V, Y, Cb, Cr)
% Input:
% R: Red, G: Green, B: Blue
% H: Hue, S: Saturation, V: Value
% Y: Luminance, Cb/Cr: Chrominance color value
% Output:
% M: Final model
% Brief description:
% This function is used to construct bounding
% rules for skin regions detection based on
% four color space models:
% RGB_1, RGB_2, HSV and YCbCr.%
%%%%%%%%%%%%%%%%%%%%%%%%%%%%%%%%%%%%%%%%%%%%%%%%%%%%%%
%% RGB_1 model %%
    RGB_1 = ((R>60) AND (G>40) AND (B>20) AND (R>G) AND...
            (R>B) AND (10<|R-G|<45) AND (|R-B|<|R-G|)) OR...
            ((|R-G|<45) AND (|R-B|>10) AND (|R-G|<|R-B|));
%% RGB_2 model %%
    RGB_2 = (0.36<=(R/(R+G+B))<=0.44) AND...
            (0.2<=(G/(R+G+B))<=0.36);
%% HSV model %%
    HSV = (0<=H<=1) AND (0.1<=S<=0.3 )AND (0.2<=V<=0.8);
%% YCbCr model %%
    YCbCr = (110.5<=Cb<=135.50) AND (135<=Cr<=145);
%% Final model %%
    M = (RGB_1) AND (RGB_2) AND (HSV) AND (YCbCr);
%%%%%%%%%%%%%%%%%%%%%%%%%%%%%%%%%%%%%%%%%%%%%%%%%%%%%%%%
```

Figure 4.4: Algorithm to detect skin regions

.

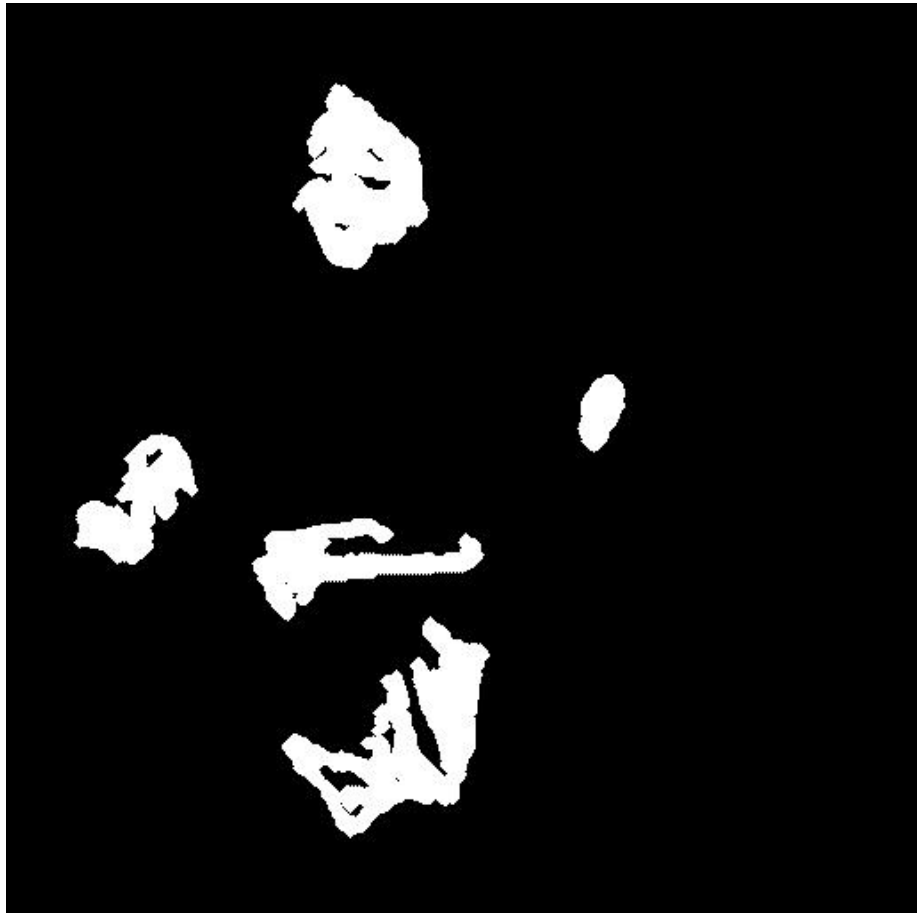

Figure 4.5: Skin region detected using color models. White region represents skin.

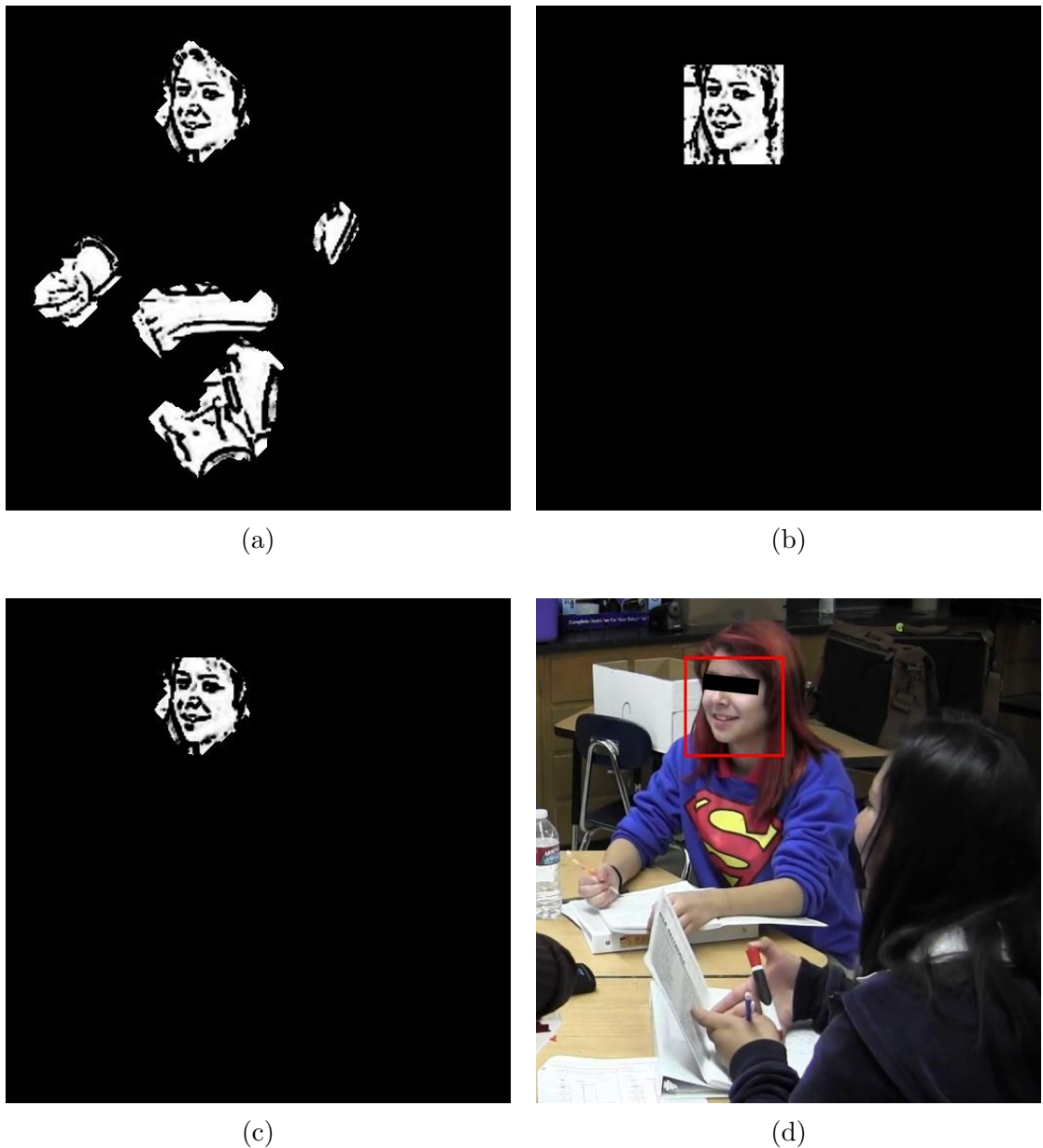

Figure 4.6: Face detection: (a) Skin areas in phase component. (b) Face detection with KNN classifier. (c) Same area of Fig.4.6a and Fig. 4.6b. (d) Final face detection result.

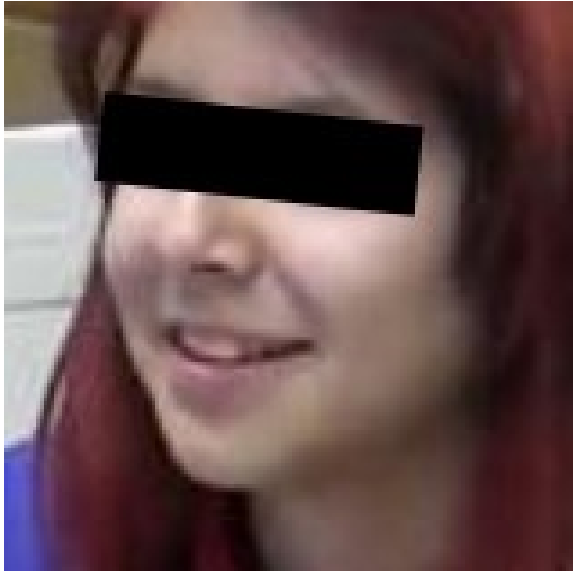

Figure 4.7: A block of face

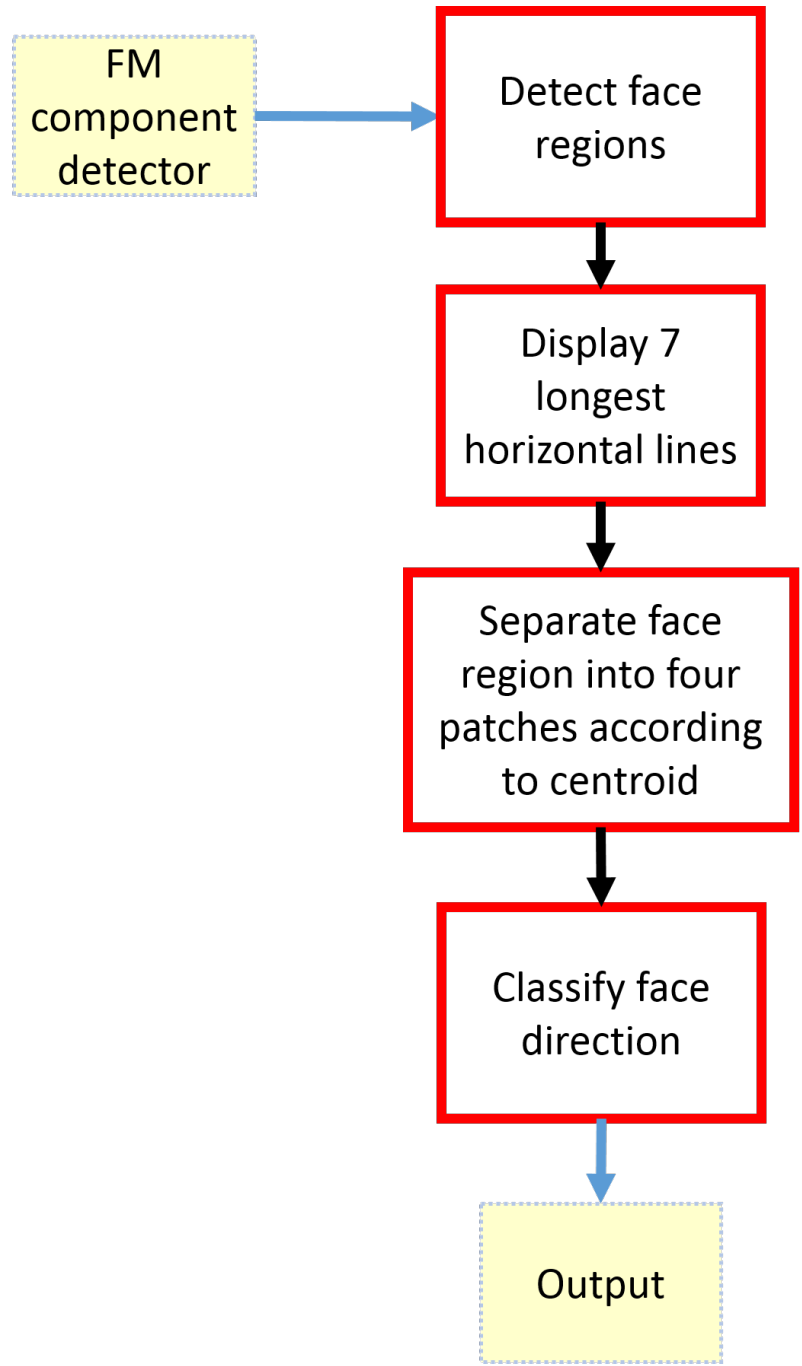

Figure 4.8: Method for face direction detection.

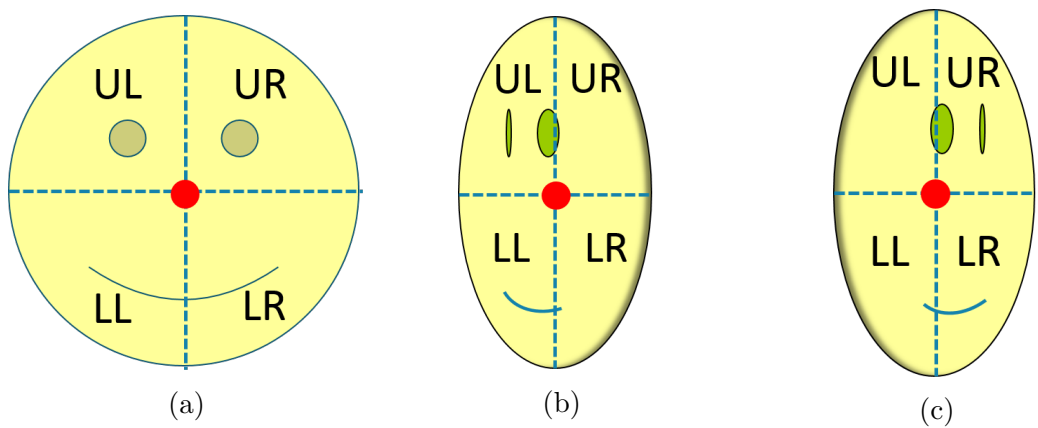

Figure 4.9: Attention classification: (a) Face image looking towards the camera. (b) Face image looking to the left. (c) Face image looking to the right.

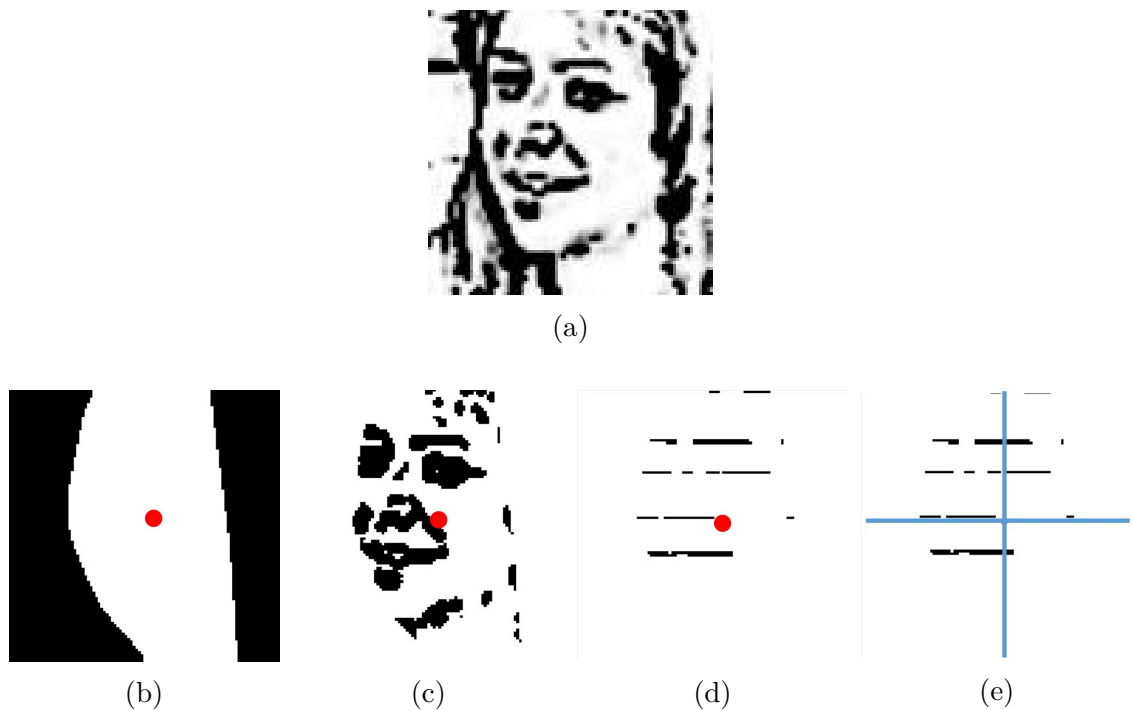

Figure 4.10: Face direction detection: (a) FM image. (b) Convex hull of the largest connected component. The red dot denotes the estimated centroid. (c) Extracted face region around the centroid. (d) Top 7 rows with the largest numbers of pixels extracted from (c). The four patches are also shown (UL, UR, LL, LR).

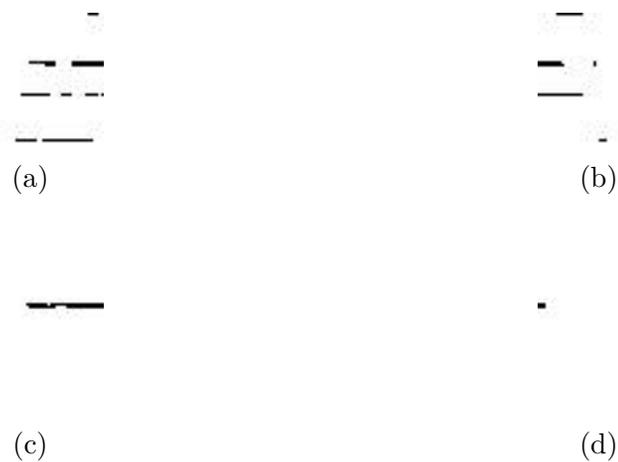

Figure 4.11: Face decomposition into four patches: (a) Upper-Left patch. (b) Upper-Right patch. (c) Lower-Left patch. (d) Lower-Right patch.

```
%%%%%%%%%%%%%%%%%%%%%%%%%%%%%%%%%%%%%%%%%%%%%%%%%%%%%%%%%%%%%%%
function ClassifyFaceDirection (UL, LL, UR, LR)
% Input:
% UL: the number of black pixels in upper-left face
% LL: the number of black pixels in lower-left face
% UR: the number of black pixels in upper-right face
% LR: the number of black pixels in lower-right face
% Brief description:
% This function is used to classify face direction
%%%%%%%%%%%%%%%%%%%%%%%%%%%%%%%%%%%%%%%%%%%%%%%%%%%%%%%%%%%%%%%
    if UL >= UR then
        if LL >= LR then
           Classify Left
        else if (UL + LL) >= (UR + LR) then
           Classify Right
        else
           Classify Left
        end if
    else if LL >= LR then
        if (UL + LL) >= (UR + LR) then
           Classify Left
        else
           Classify Right
        end if
    end if
%%%%%%%%%%%%%%%%%%%%%%%%%%%%%%%%%%%%%%%%%%%%%%%%%%%%%%%%%%%%%%%
```

Figure 4.12: Algorithm to classify face direction

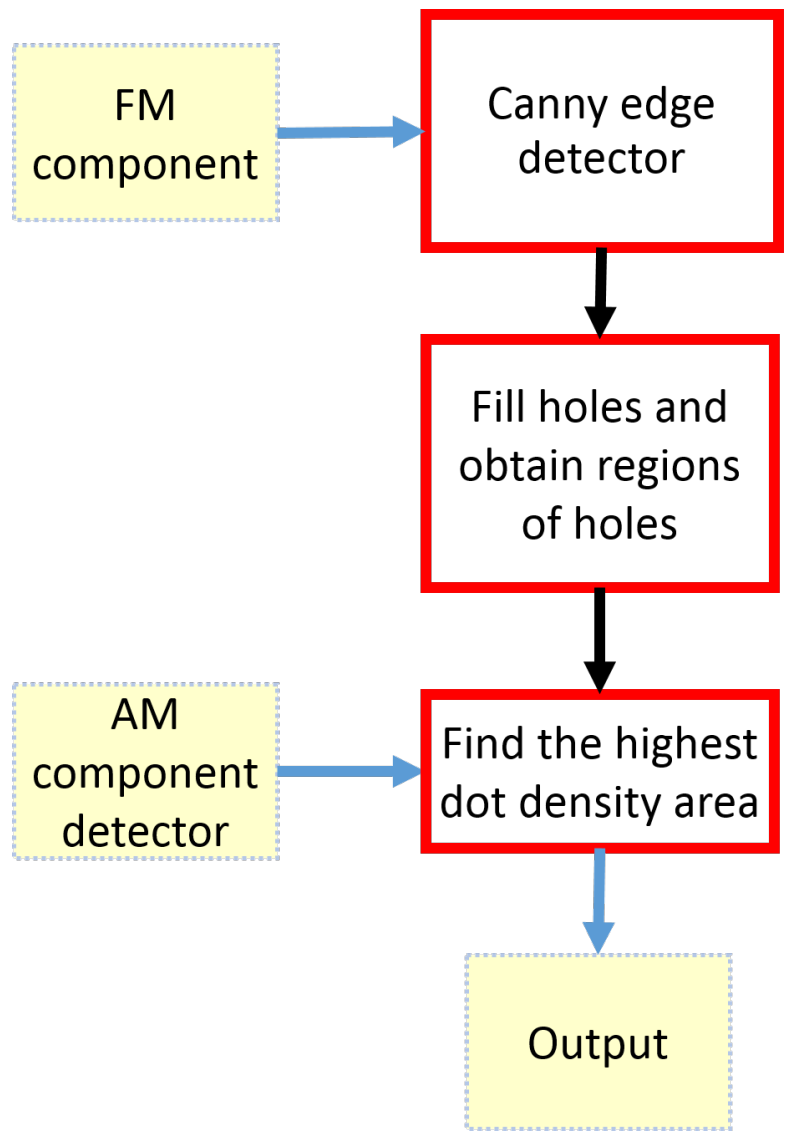

Figure 4.13: Method of back of the head detection

```
%%%%%%%%%%%%%%%%%%%%%%%%%%%%%%%%%%%%%%%%%%%%%%%%%%%%%%%%%%%%%%%%%%%%%%%
function FindTheHighestDotDensityArea (I, m, n)
% Input:
% I: an input binary image
% m: the height of I
% n: the width of I
% Output:
% H: the block that has the highest dot density
% Brief description:
% This function is used to find the highest
% dot density area. The size of this area is 200px*200px.
%%%%%%%%%%%%%%%%%%%%%%%%%%%%%%%%%%%%%%%%%%%%%%%%%%%%%%%%%%%%%%%%%%%%%%%
    Set s = 200;
    Set Rate = a empty matrix;
    for 0 < i < m-s+1
        for 0 <j <n-s+1
            Num = the number of black pixels in...
            the area from I(i,j) to I(i+199, j+199);
            Rate(i, j) = Num / (200 * 200);
        end
    end
    Max_Rate = the max number in matrix 'Rate';
    Max_i = the 'i' value when Rate = Max_Rate;
    Max_j = the 'j' value when Rate = Max_Rate;
    H = the area from I(Max_i, Max_j) to I(Max_i+199, Max_j+199);
%%%%%%%%%%%%%%%%%%%%%%%%%%%%%%%%%%%%%%%%%%%%%%%%%%%%%%%%%%%%%%%%%%%%%%%
```

Figure 4.14: Algorithm to find the highest density area of black pixels

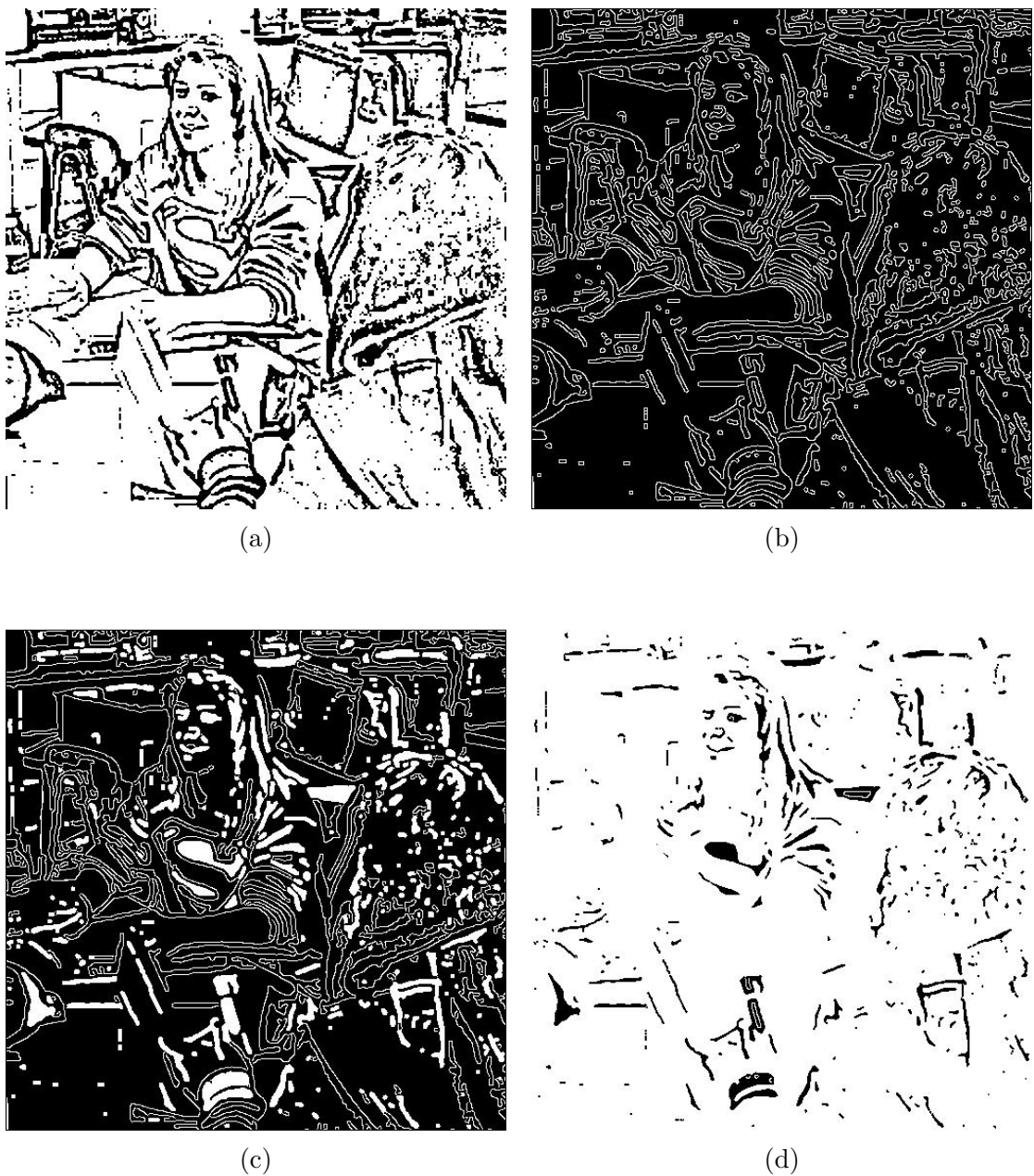

(a)

(b)

(c)

(d)

Figure 4.15: Back of the head detection: (a) Binary image of FM component. (b) Edge detection with Canny approximate. (c) Filling holes in Fig. 4.15b. (d) The complement of filled holes.

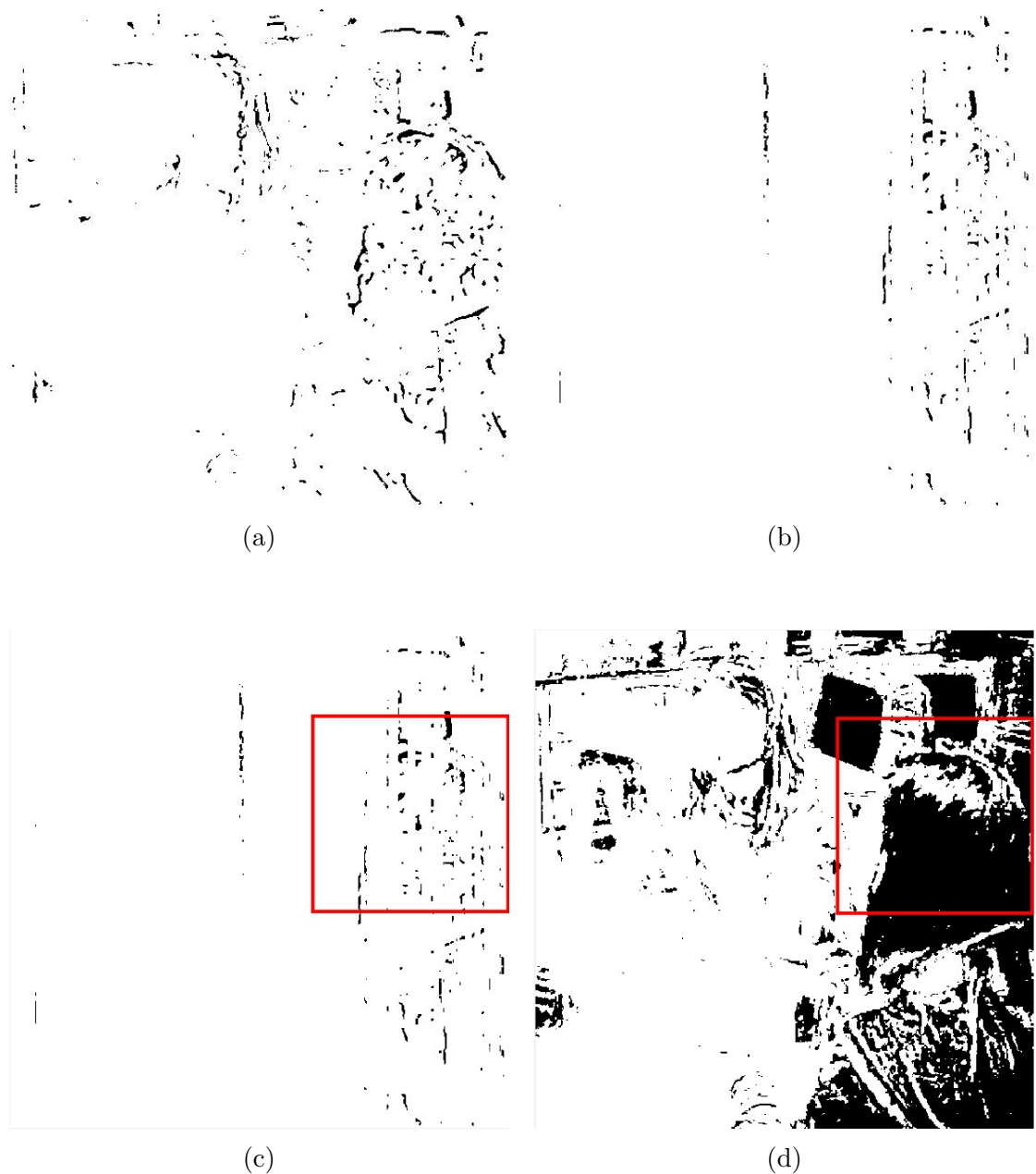

Figure 4.16: Back of the head detection: (a) Doing 'AND' operation between Fig. 4.15d and binary image of AM component. (b) 60 longest vertical lines. (c) The highest density area of black pixels. (d) Same area in AM component.

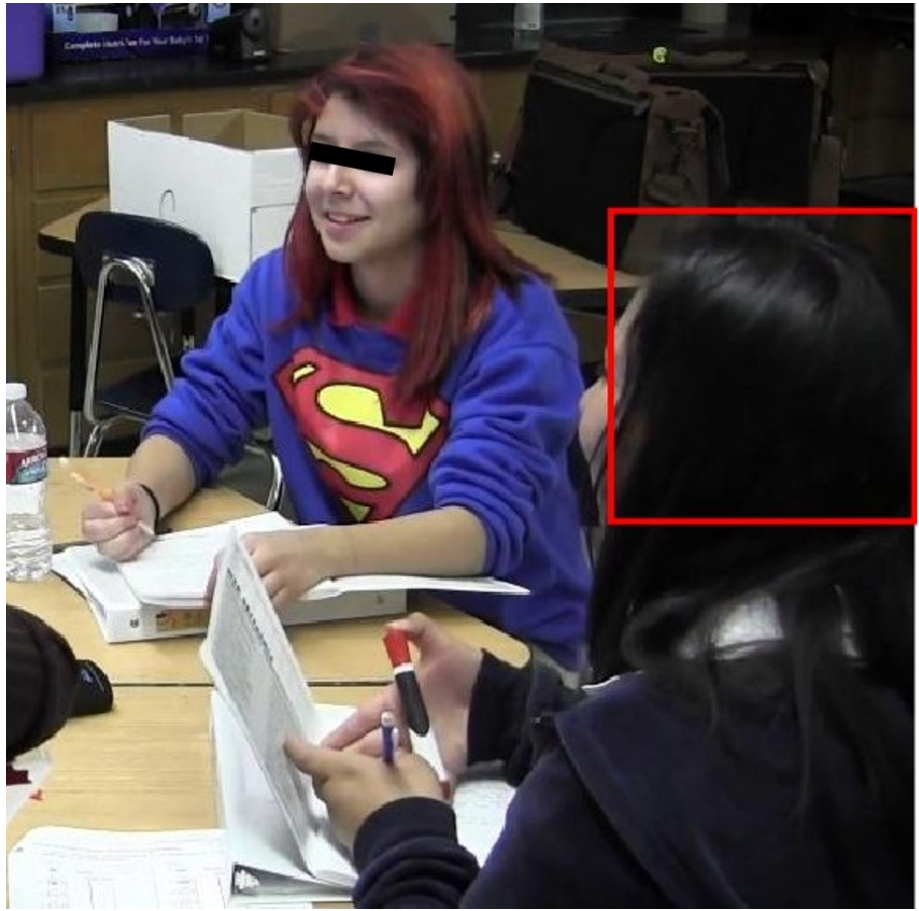

Figure 4.17: Back of head detection result

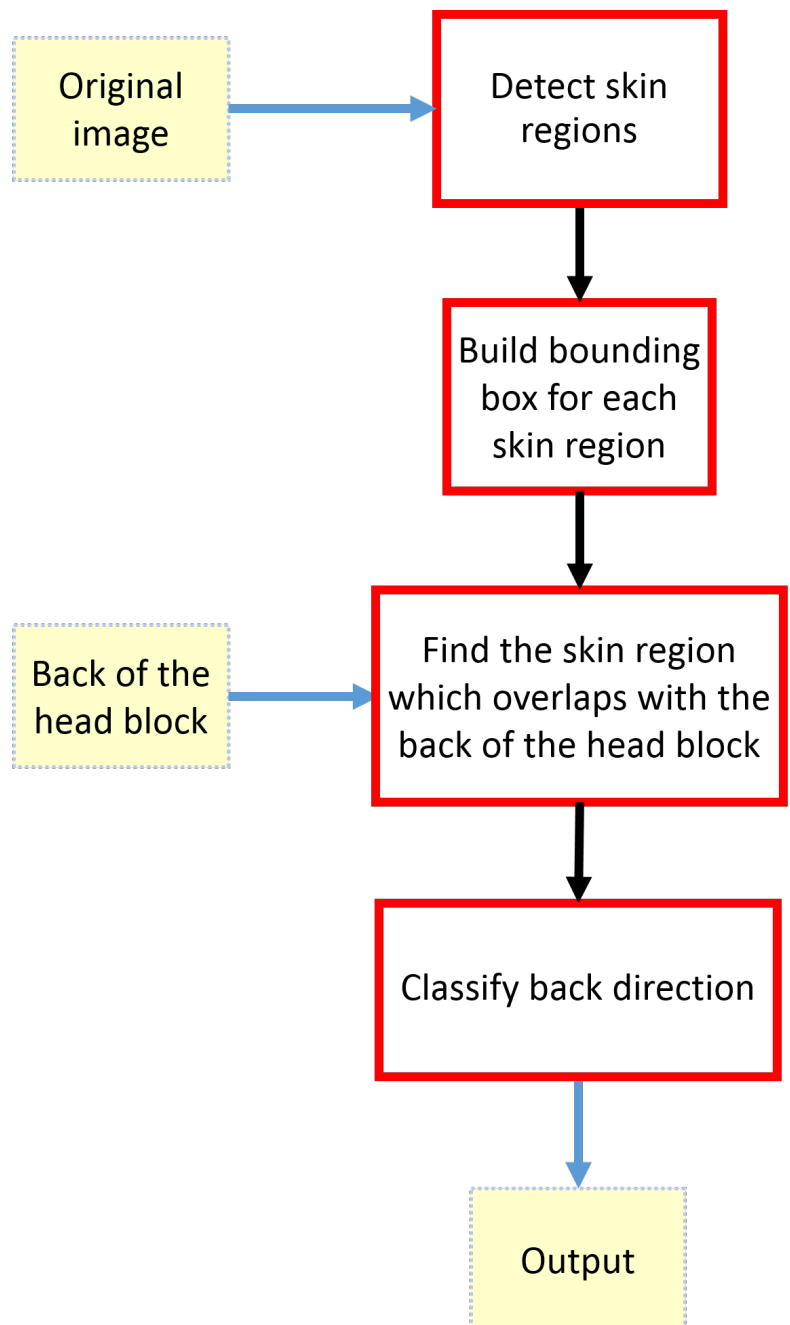

Figure 4.18: Method for back of the head direction detection.

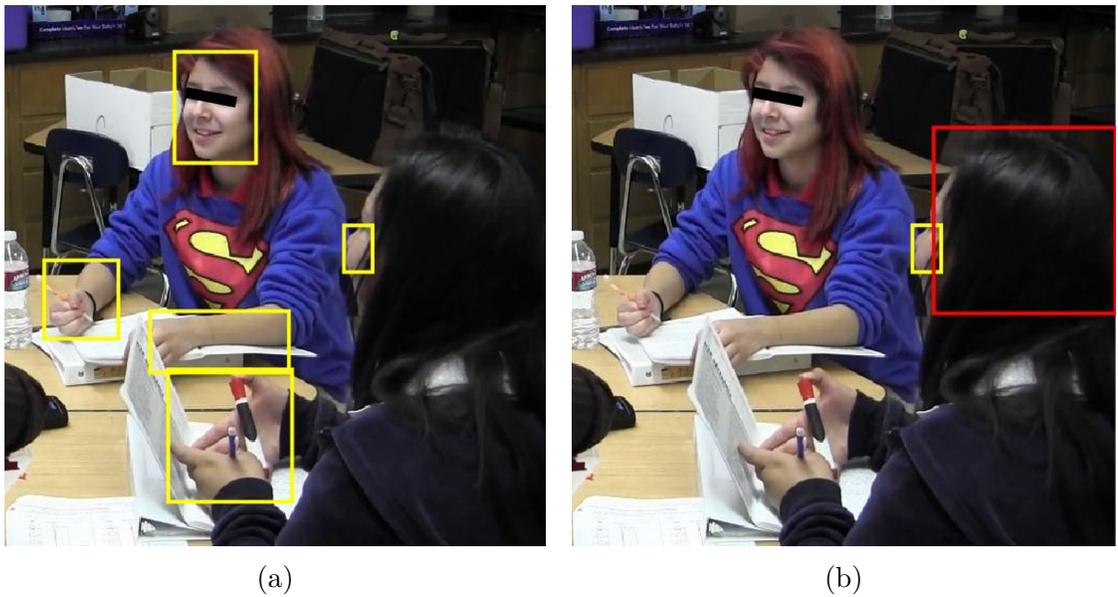

(a)                   (b)

Figure 4.19: Back of the head detection: (a) Bounding box for all skin regions. (b) Final face and back of head position.

# Chapter 5

# Results

The datasets for this thesis were taken from the Advancing Out-of-school Learning in Mathematics and Engineering (AOLME) project videos. More than 8,000 frames were extracted from over 25 short different video segments. The videos contain students' activities in unrestrained environments. Among other activities, the videos captured students talking, listening, writing, and typing. In these videos, students sit around the table at random. Some of the students are recorded facing towards the camera while others are looking away from the camera. For the students facing away from the camera, we need to use the back of the head detector.

## 5.1 Face Direction Results for Students Looking towards the Camera

For face direction detection, the dataset consisted of 7,111 images of students looking to the left (from ten students), and 6,154 images of students looking to the right (from five students). Some of the extracted FM components are shown in Figs. 5.1 5.2 provide several FM component samples of these faces.

We present the classification results in Figs. 5.3, 5.4 and 5.5. For each patch classifier (low, upper, and whole face), images that generate points above $y = x$ are classified as looking to the right. Similarly, images that generate points below $y = x$ are classified as looking to the left. The final classification is based on the majority of the patch classification results. Thus, as an example, an image is classified as looking left if two or all three of the patch classifiers returned a left classification. The final results are presented in Table 5.1. The looking-left classifier had an 97.1% correct classification rate and the looking-right classifier gave 95.9% correct classification rate.

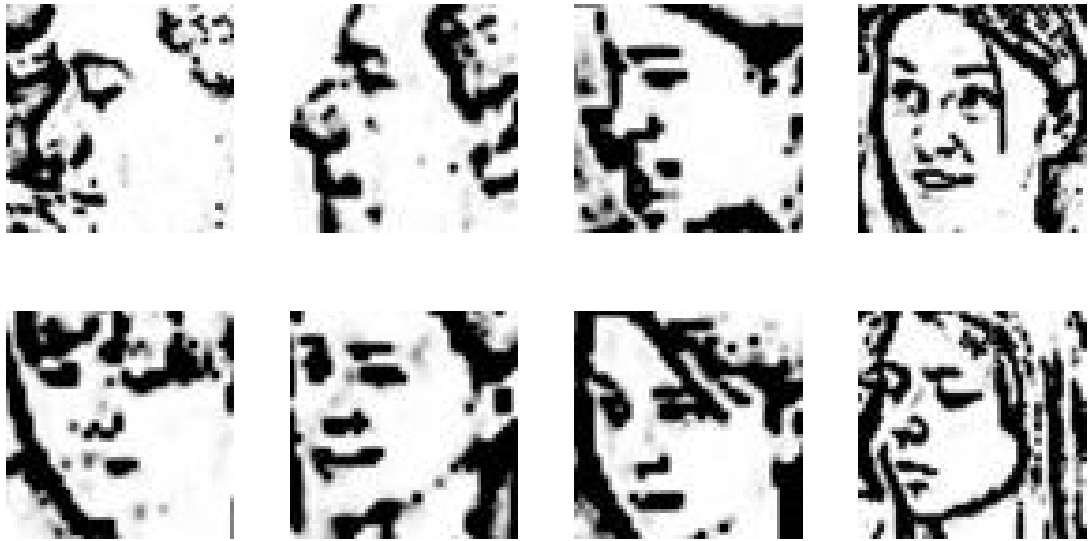

Figure 5.1: Face images for students and facilitator looking to the left

We also present representative classification results. Fig. 5.6 shows the FM components of cases that were correctly classified. In these example, facial features are hard to see due to variability in the image acquisition. Nevertheless, the proposed approach was sufficiently robust to avoid these issues.

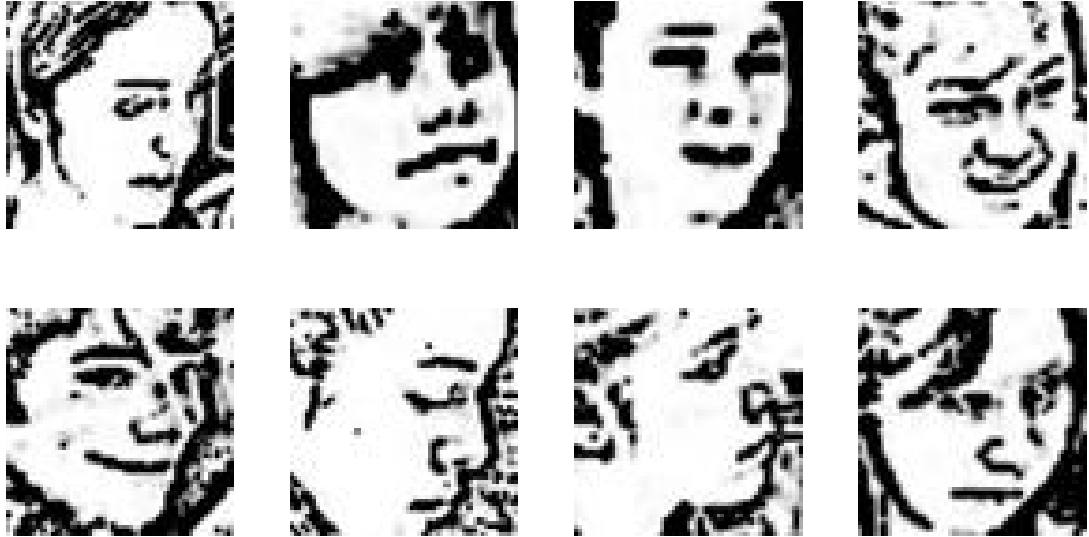

Figure 5.2: Face images for students and facilitator looking to the right

Table 5.1:  Classification Results for Face Direction Detection.

| Classifier | Number of images | Accuracy |
|------------|------------------|----------|
| Left       | 7111             | 97.1 %   |
| Right      | 6154             | 95.9 %   |

## 5.2   Results for images of students looking away from the camera

For classifying images where students are looking away from the camera, the dataset consisted of 1,718 left-looking images and 4,404 right-looking images from 5 people. Figs. 5.7 and 5.8 provide several image samples. The results are summarized in Table 5.2. The looking-left classifier had an 87.6% correct classification rate and the looking-right classifier gave 93.3% correct classification rate.   A difficult example

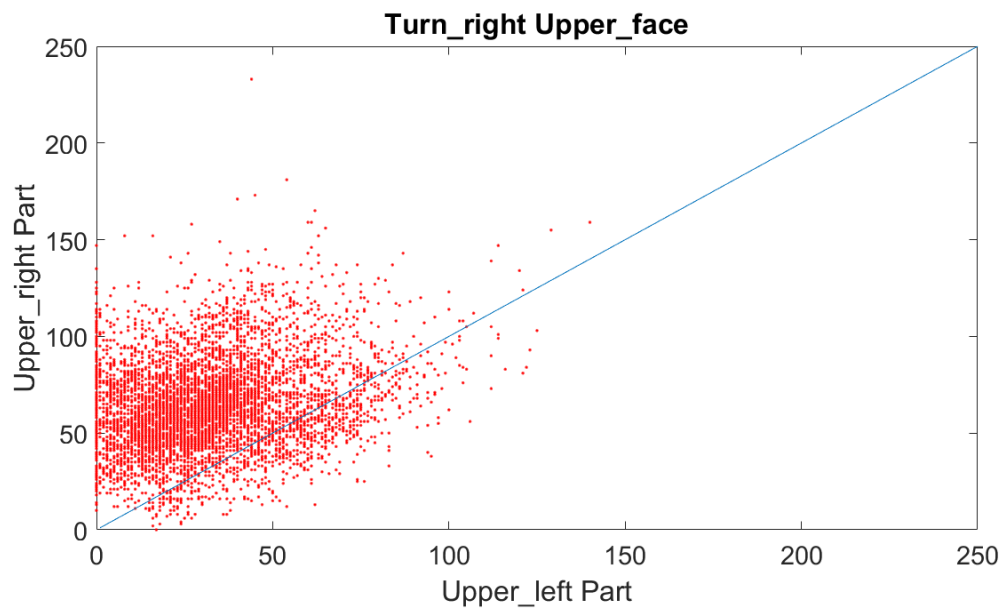

(a)

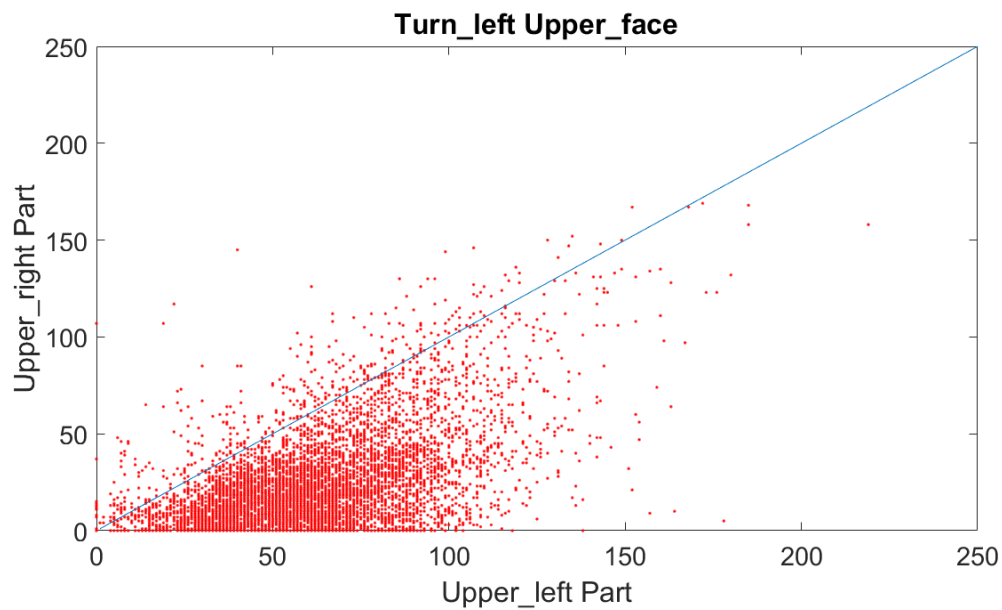

(b)

Figure 5.3: Plots for: (a) Upper face classifier for looking right images. (b) Upper face classifier for looking left images.

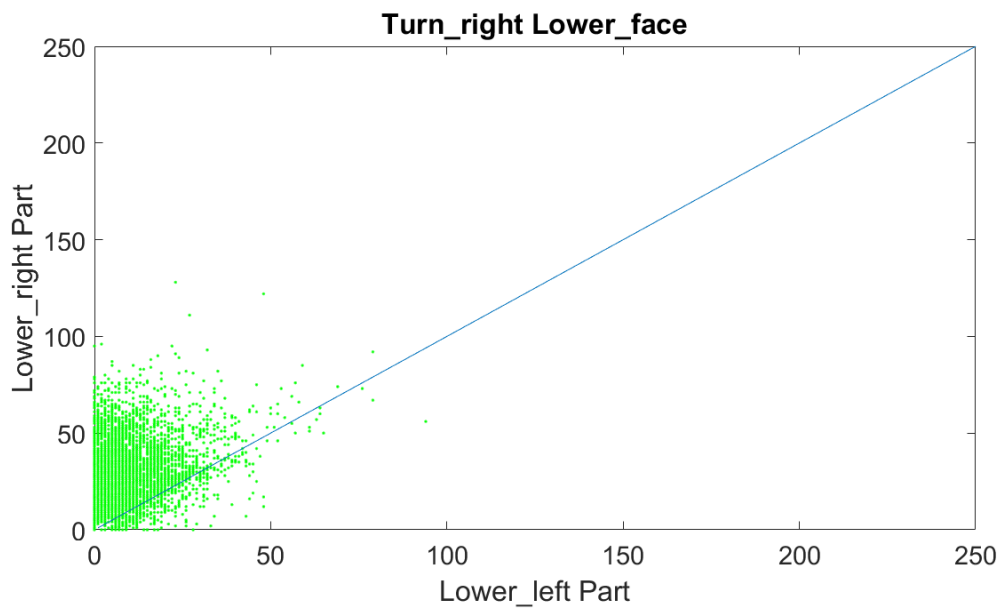

(a)

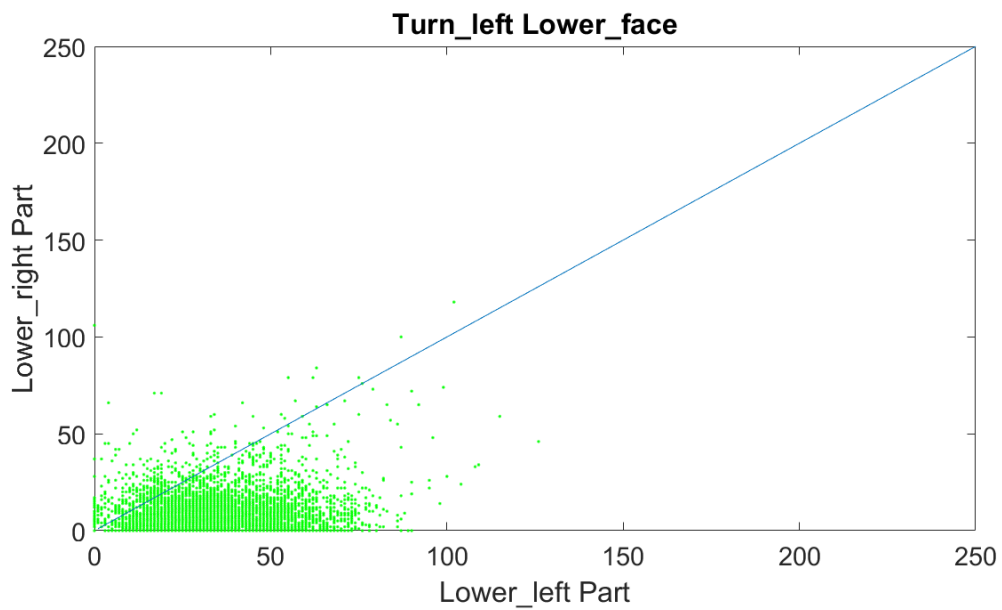

(b)

Figure 5.4: Plots for: (a) Lower face classifier for looking right images. (b) Upper face classifier for looking left images.

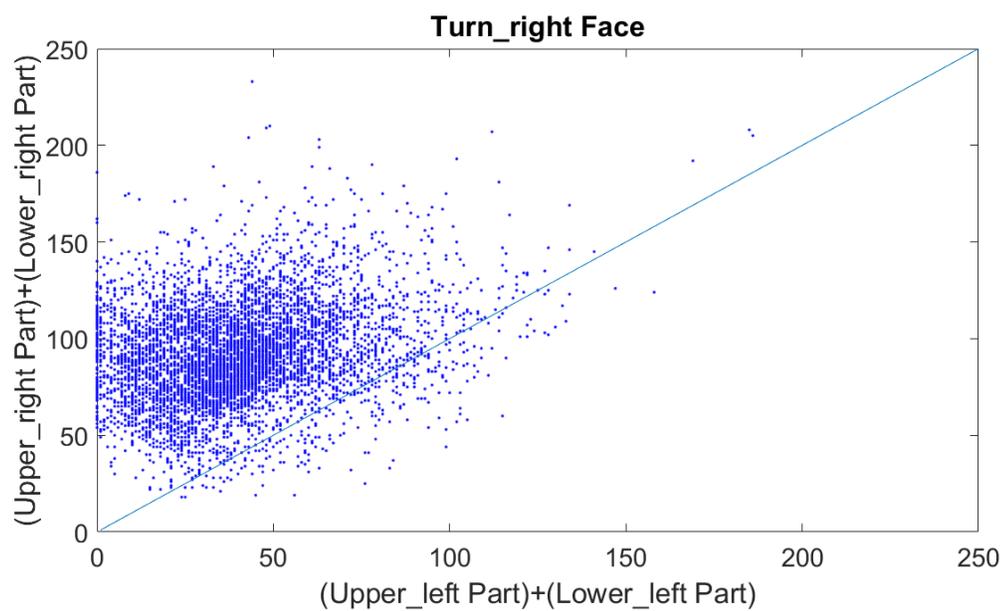

(a)

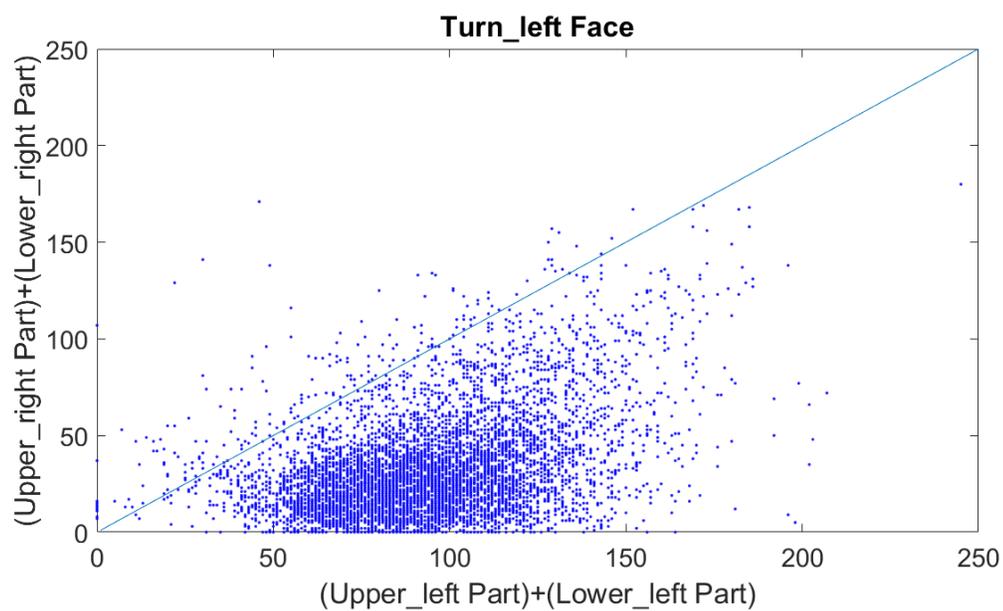

(b)

Figure 5.5: Plots for: (a) Whole face classifier for looking right images. (b) Whole face classifier for looking right images.

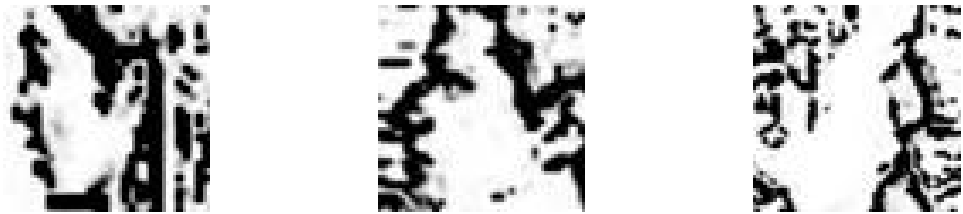

Figure 5.6: Successful face direction detection for difficult cases.

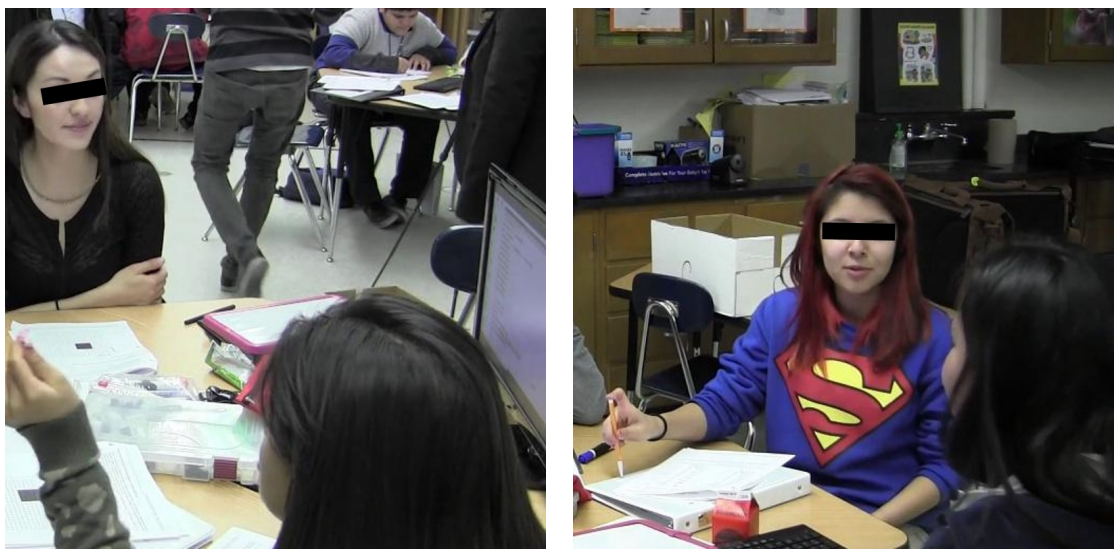

Figure 5.7: Two left samples for back of the heads

is shown in Fig. 5.9. In this case, the head and the sweater shared the same color and similar texture characteristic. Thankfully, the dot density detector was able to differentiate between the two and thus allowed the system to correctly the classify the input image.

Table 5.2: Classification Results for Back of the Head Direction Detection

| Classifier | Number of images | Accuracy |
|---|---|---|
| Left | 1718 | 87.6 % |
| Right | 4404 | 93.3 % |

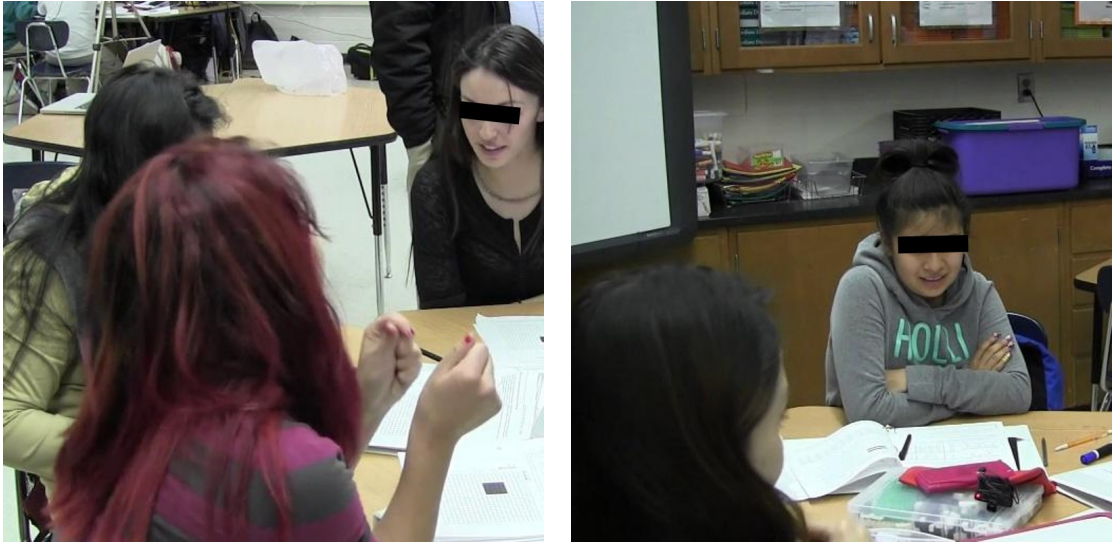

Figure 5.8: Two back of the head examples that were correctly classified.

## 5.3  Failed Classification Examples

We present three examples that were not classified correctly. One example for students looking towards the camera and another two for students looking away from the camera.

We present an example that was incorrectly classified in Fig. 5.10 . In this case, the girl's bangs interfered with the classification results. The bangs produced extra pixels in the upper left part of the face resulted in a looking-left as opposed to the correct, looking right classification.

We present two examples for students looking away from the camera in Fig. 5.11. In one example, the back of the head is combined with the student's hand. In another

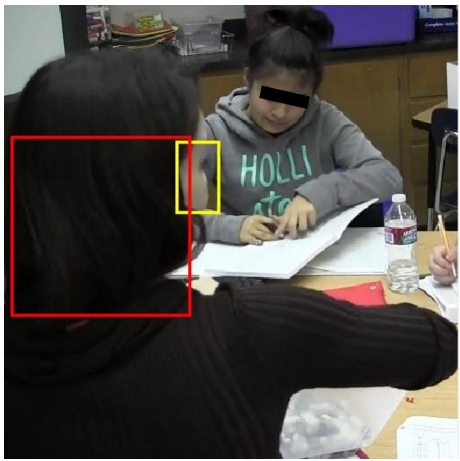

(a)

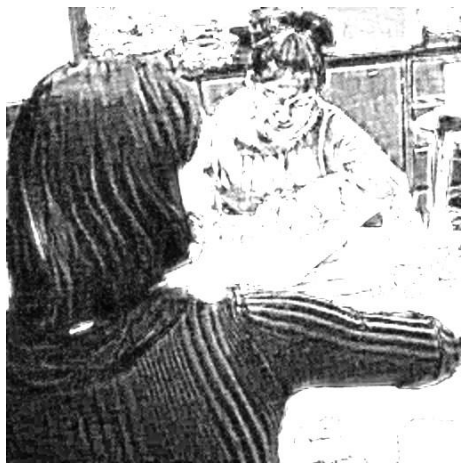

(b)

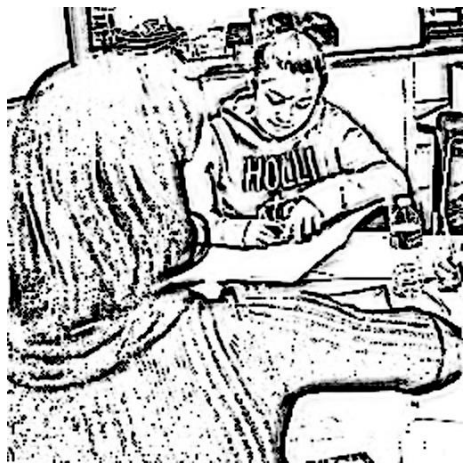

(c)

Figure 5.9: Back of the head direction detection for a difficult case: (a) Original image. (b) AM component. (c) FM component.

example, a hand was combined with another face.

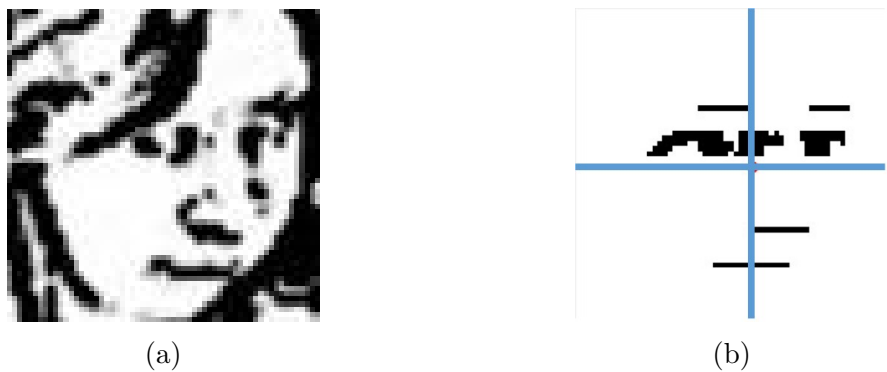

(a)                                    (b)

Figure 5.10: A failed example: (a) Phase component of a face. (b) Separating face features into four patches.

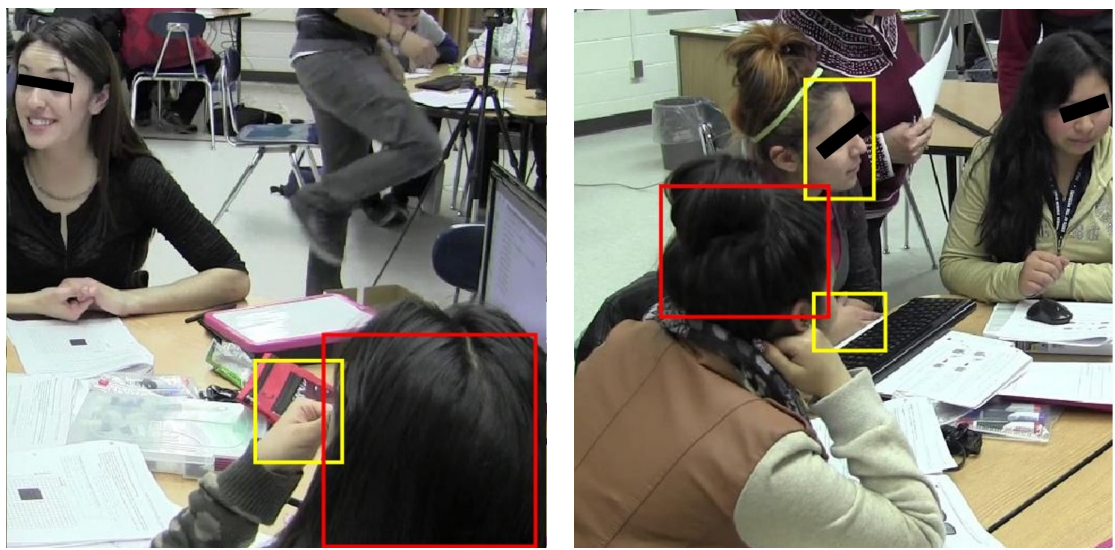

Figure 5.11: Failed examples for back of the head direction detection.

# Chapter 6

# Conclusion and Future Work

## 6.1 Conclusion

In this thesis, new methods were proposed for detecting human attention in AOLME videos. The primary contributions of the new methods included: (i) a phase-based method, (ii) a new method for detecting back of the head, (iii) a new approach for detecting whether the students were looking to the left or right. Furthermore, a new classifier was used to determine attention in students looking away from the camera. In the proposed methodology, both the extracted AM and FM components were used to detect the face and the back of the head.

The datasets were challenging because the videos were recorded in uncontrolled environments where the students in the videos were recorded from different angles, including the front and the back. The datasets consisted of $13,265$ face images from 10 people and 6122 back of the head images from 5 people. Overall, the proposed methods achieved an accuracy of 97.1% for left-looking front face images, 95.9% for right-looking front face images, 87.6% for left-looking back of the head images, and 93.3% for right-looking back of the head images. Thus, the results showed that

AM-FM based methods hold great promise for analyzing human activity videos

## 6.2   Future Work

Future work should include:

- **Exploring methods to detect the back of the heads for boys.** Due to their short hair, phase based methods will have fewer pixels to work with in this case. This makes it difficult to detect their heads from images in which they are looking away from the camera. New methods will have to be developed to cover this case.

- **Color independent back of the head detection**. When detecting the back of the head, hair color plays an important role in the AM component. This makes back of the head detection algorithm perform poorly for non-dark hair. We need to explore new methods that will work for different hair colors.

- **Extending attention detection to include other head poses.** This thesis classifies student's attention into looking right or looking left. More directions need to be considered in the future (e.g., looking up, looking down, looking sideways, etc).

- **Studies on large video databases.** Clearly, it will be interesting to extend the study to larger video databases that cover different environments and more complex human interactions.